%% file: main.tex
\documentclass[10pt,twocolumn,letterpaper]{article}

\usepackage{cvpr}              

\input{preamble}

\definecolor{cvprblue}{rgb}{0.21,0.49,0.74}
\usepackage[pagebackref,breaklinks,colorlinks,allcolors=cvprblue]{hyperref}

\def\paperID{44936} 
\def\confName{CVPR}
\def\confYear{2026}

\title{Deep Feature Deformation Weights}

\author{Richard Liu\\
University of Chicago\\
{\tt\small guanzhi@uchicago.edu}
\and
Itai Lang\\
University of Chicago\\
{\tt\small itailang@uchicago.edu}
\and 
Rana Hanocka\\
University of Chicago\\
{\tt\small ranahanocka@uchicago.edu}
}

\input{macros.tex}

\begin{document}

\twocolumn[{%
\renewcommand\twocolumn[1][]{#1}%
\maketitle
\begin{center}
    \centering
    \captionsetup{type=figure}
    \includegraphics[width=\textwidth]{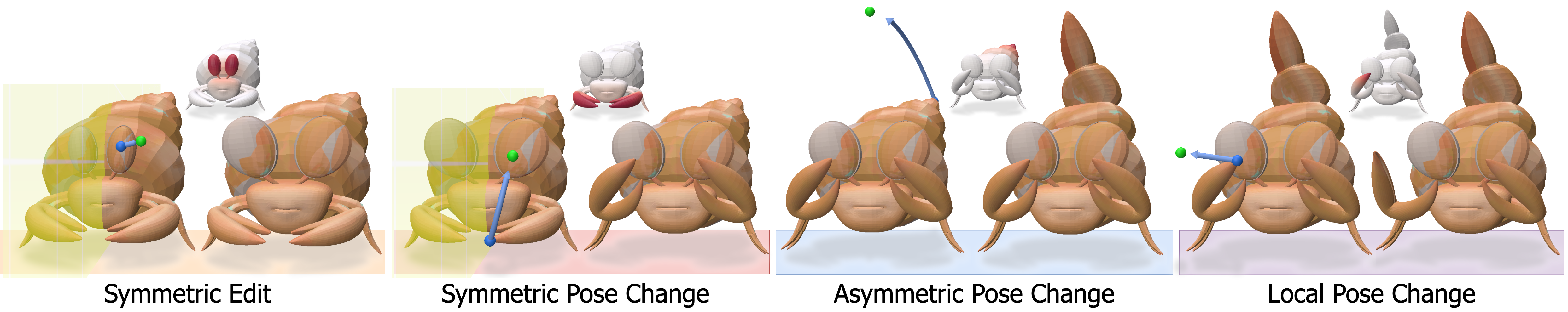}
    \captionof{figure}{Our DFD framework enables flexible control over a wide range of deformations in real time. Symmetric deformations may be achieved through our automatically detected symmetry plane (yellow).}
    \label{fig:teaser}
\end{center}
}]
\input{0_abstract}
\input{01_introduction.tex}
\input{02_related_work.tex}
\input{03_method.tex}
\input{04_experiments.tex}

\input{05_conclusion.tex}
\input{06_acknowledgements.tex}
{
    \small
    \bibliographystyle{ieeenat_fullname}
    \bibliography{references}
}

\input{supplementary/supplementary.tex}

\end{document}

%% file: preamble.tex
\usepackage{array}
\usepackage{multirow}
\usepackage{multicol}
\usepackage{pifont}
\usepackage[dvipsnames]{xcolor}
\usepackage{booktabs}
\usepackage{overpic}

\newif\ifdraft
\draftfalse

\ifdraft
\newcommand{\rone}[1]{{\color{BurntOrange}[\textbf{3dFw:} #1]}}
\newcommand{\rtwo}[1]{{\color{ForestGreen}[\textbf{8NZN:} #1]}}
\newcommand{\rthree}[1]{{\color{Plum}[\textbf{NfHn:} #1]}}

\else
\newcommand{\rone}[1]{}
\newcommand{\rtwo}[1]{}
\newcommand{\rthree}[1]{}

\fi



%% file: macros.tex
\newif\ifdraft
\draftfalse

\ifdraft
\newcommand{\richard}[1]{{\color{orange}[\textbf{Richard:} #1]}}
\newcommand{\itai}[1]{{\color{magenta}[\textbf{Itai:} #1]}}
\newcommand{\rana}[1]{{\color{cyan}[\textbf{Rana:} #1]}}

\else
\newcommand{\richard}[1]{}
\newcommand{\itai}[1]{}
\newcommand{\rana}[1]{}

\fi

\newcommand{\ourmethod}{DFD}

\makeatletter
\DeclareRobustCommand\onedot{\futurelet\@let@token\@onedot}
\def\@onedot{\ifx\@let@token.\else.\null\fi\xspace}

\makeatother

\let\titleold\title
\renewcommand{\title}[1]{\titleold{#1}\newcommand{\thetitle}{#1}}
\def\maketitlesupplementary
   {
   \newpage
       \twocolumn[
        \centering
        \Large
        \textbf{\thetitle}\\
        \vspace{0.5em}Supplementary Material \\
        \vspace{1.0em}
       ] 
   }

\AtEndPreamble{
    \usepackage[capitalize]{cleveref}
    \crefname{section}{Sec.}{Secs.}
    \Crefname{section}{Section}{Sections}
    \Crefname{table}{Table}{Tables}
    \crefname{table}{Tab.}{Tabs.}
}

\RequirePackage{enumitem}
\setlist[itemize]{noitemsep,leftmargin=*,topsep=0em}
\setlist[enumerate]{noitemsep,leftmargin=*,topsep=0em}

%% file: 0_abstract.tex
\begin{abstract}
Handle-based mesh deformation is a classic paradigm in computer graphics which enables intuitive edits from sparse controls. Classical techniques are fast and precise, but require users to know ideal handle placement apriori, which can be unintuitive and inconsistent. Handle sets cannot be adjusted easily, as weights are typically optimized through energies defined by the handles. Modern data-driven methods, on the other hand, provide semantic edits but sacrifice fine-grained control and speed. We propose a technique that achieves the best of both worlds: deep feature proximity yields smooth, visual-aware deformation weights with no additional regularization. Importantly, these weights are computed in real-time for any surface point, unlike prior methods which require expensive optimization. We introduce barycentric feature distillation, an improved feature distillation pipeline which leverages the full visual signal from shape renders to make distillation complexity robust to mesh resolution. This enables high resolution meshes to be processed in minutes versus potentially hours for prior methods. We preserve and extend classical properties through feature space constraints and locality weighting. Our field representation enables automatic visual symmetry detection, which we use to produce symmetry-preserving deformations. We show a proof-of-concept application which can produce deformations for meshes up to 1 million faces in real-time on a consumer-grade machine. Project page at \url{https://threedle.github.io/dfd}.

\end{abstract}

%% file: 01_introduction.tex
\section{Introduction} \label{sec:introduction}
Handle-based deformation frameworks enable intuitive editing with sparse inputs. Traditional methods solve an optimization problem to obtain either a weight matrix for linear blending of handles \cite{BBW2011, wang_linear_2015} or the deformed mesh vertices directly \cite{ARAP_modeling:2007}. Both types of methods require strategic placement of the handles to obtain desirable deformations~\cite{kim2023optctrlpoints}. Traditionally, local influence of the handles is enforced through the optimized energy (typically a Laplacian or rigidity energy). Local influence is assumed to be desirable, but in this work we argue that \emph{global/semantic influence} can also be desirable (e.g. co-deformation of chair legs). In fact, prior work provides evidence for this \cite{10.1145/1531326.1531339, wang_3dn_2019, liu2021deepmetahandles, yoo2024plausible}. These data-driven methods offer semantic-aware edits (symmetry/structure preserving), but lack the fine-grained control and speed of traditional methods. 

Our work aims to synthesize the strengths of the these competing approaches. Specifically, we desire the speed and fine-grained control of traditional handle-based deformation methods, while simultaneously capturing visual understanding from data priors. This brings us to \emph{Deep Feature Deformation weights} (DFD weights), which use distilled deep features from pretrained 2D models to define the function mapping handle transformations to surface deformations. Importantly, this function is \emph{not} conditioned on the choice of handle set and does not involve solving any optimization problem. We take a simple yet effective approach: we define linear blending weights based on feature similarities. We demonstrate that these weights are robust across deformation types, handle choice, and shape.

We specifically propose to parameterize the space of weights through a neural field, so that \emph{any} handle can be chosen from ambient space, without the need for expensive re-optimization. DFD weights correlate visually similar structures due to the data prior of 2D pre-trained models. Deformations using our weights are smooth without requiring any additional vertex constraints or regularization. 

Each field is optimized per-shape, but distillation is fast and accelerated through a novel \textit{barycentric feature distillation}. This procedure allows feature fields to be rapidly distilled on coarse shapes and transferred to high-resolution counterparts during inference. Existing distillation techniques distill to the mesh resolution. We instead make distillation a function of \emph{render resolution}, and leverage the geometric prior of the mesh to efficiently sample the shape space from each render. We show that with barycentric distillation, feature fields for shapes ranging from 1000 to $>$10 million faces can all be optimized within a few minutes.

Though our weights by default give global deformations, we still account for locality and fixed point constraints. Specifically, we introduce locality through a geodesic-weighting adjustment. Under our framework, point constraints naturally extend to visual constraints, which we dub \textit{feature space constraints.} Visually similar parts constrained by these fixed points are preserved under deformation.

We evaluate our DFD weights on shapes of varying quality, resolution, and type. Our weights are robust to topological deficiencies and our performance is on-par or outperforms all baselines on their specialized datasets.  We develop a toy GUI for interactive mesh deformation.

%% file: 02_related_work.tex
\section{Related Work} \label{sec:related_work}

We divide related handle-based deformation work into traditional methods and data-driven ones. We then address a set of common desirable properties of such methods, propose some alterations, and contextualize our method against relevant baselines in \cref{tbl:properties}. 

\subsection{Traditional Handle-Based Methods}
Handle-based methods offer low-dimensional control structures for performing shape editing or animation tasks. These methods can be roughly categorized by their choice of control structure, the most common being handle points, cages, or skeletal rigs. 

Methods which use handle sets either use variational methods to optimize each deformation according to target handle positions (\citep{ARAP_modeling:2007, 10.1145/1531326.1531340, 4359478}, or optimize a weight matrix to perform linear blending of the prescribed handle positions to produce the deformed shape (\citep{wang_linear_2015,BBW2011}).  Implicit-ARAP~\cite{baieri2025implicit} is an interesting recent work which extends as-rigid-as-possible optimization to implicit representations. 

Cages~\cite{lipman2008green, ju2023mean, joshi2007harmonic, hormann2008maximum, weber2011complex} are similar in that they define control handles which form the nodes of a coarse polytope enclosing the mesh. Each vertex is defined as a linear combination of node positions, the weights of which are typically generalized barycentric coordinates \cite{Floater_2015}.

Skeleton rigs define a hierarchy of bones related through joint rotations, and deformations applied to the bones are transferred to the mesh surface through skinning weights~\cite{weber2007context, baran2007automatic, kavan2007skinning, magnenat1989joint, le2016real}. The hierarchical structure allows for movements made to the extremities to propagate through the kinematic chain, causing larger scale pose changes and shape motions. This property is what makes skeleton rigs preferred in animation and shape-preserving applications.  

All traditional methods aim to produce some form of ``natural deformation'' \cite{4359478}, where in lieu of a well-defined mathematical notion of ``natural'', distortion minimization becomes the stand-in objective. Distortion minimization is an effective approach to achieving ``pose change'' types of deformations, where articulated parts are bent or twisted while preserving total shape volume, but these are not the only types of deformations a user may desire. For instance, a user may wish to elongate all the legs of a chair mesh in a symmetric way to preserve the chair structure and function. Data-driven approaches have since emerged to champion these types of surface-distorting, yet desirable, edits. 

\subsection{Data-Driven Handle-Based Methods}
Early data-driven shape deformation works explored the space of semantic design attributes through curated datasets \cite{10.1145/2766908}. Recent works use data to predict the parameters of the traditional control structures cited in the previous section. Neural Cages~\cite{yifan2020neural} introduces a neural network which predicts both the cage node positions and the corresponding node translations to deform a shape towards a target. Similar works have utilized shape targets for optimizing control structure quantities, such as AlignNet~\cite{hanocka2018alignet}, OptCtrlPoints~\cite{kim2023optctrlpoints}, KeypointDeformer~\cite{jakab2021keypointdeformer}, DeepMetaHandles~\cite{liu2021deepmetahandles}, and DeformSyncNet \cite{10.1145/3414685.3417783}. NeuralMLS~\cite{shechter2022neuralmls} leverages the smooth prior of neural networks to obtain deformation weights based on input handles.

APAP~\cite{yoo2024plausible} is a recent method which combines the classical approach of ARAP with the modern approach of supervision from pretrained text-to-image models to generate "plausibility-aware" shape deformations from handle positions and anchor points. 

Importantly by learning through data, these works demonstrate user-desirable deformations which are not necessarily near-isometric pose changes. In particular, IWires \cite{10.1145/1531326.1531339}, DeepMetaHandles \cite{liu2021deepmetahandles}, and APAP \cite{yoo2024plausible} show surface distorting transformations (e.g. part scaling and stretching) that preserve global symmetries and shape structure. Our method aligns with this line of work, though we emphasize that our method is ultimately capable of both types of edits, as demonstrated in Fig.~\ref{fig:teaser}.

\subsection{Properties of Handle-Based Methods}
Most methods have abided by a set of desirable properties for handle-based deformation. As outlined by OptCtrlPoints~\cite{kim2023optctrlpoints}, these properties are: 
\begin{enumerate}
    \item Identity: The original shape must be reconstructed under zero
movement of shape handles.
    \item Locality: The deformation produced by each individual handle
must be local and smooth.
    \item Closed-form: The deformed shape is a closed-form expression of the target point transformations
    \item Flexibility: The deformation handles and function is defined without any additional information about the
shape (e.g. skeleton hierarchy).
\end{enumerate}
We agree with properties 1,3,4 but argue that locality is not essential. Rather, deformations may be global as long as they are smooth. Locality may be preferred depending on the user's intent, but allowing global deformations opens up the possibility of modeling with \emph{semantics} -- editing while preserving shape symmetries (visual/intrinsic/extrinsic) and global structure/function. As explained in the previous section, we are not the first work to demonstrate that such deformations could be desirable.

We propose two more desiderata: efficient compute under changing handles and mesh resolution. All cited works require solving an optimization problem to obtain either the weights or the final deformation. These works also scale poorly -- at best quadratically with vertex count. This is a fundamental bottleneck towards real-time deformation, as users will frequently iterate over handles. OptCtrlPoints~\cite{kim2023optctrlpoints} takes an important step towards reducing the bottleneck, but does not resolve the underlying issues of optimization and scaling. Our method makes handle-based recomputation fast by simply assigning weights in terms of feature distances, bypassing the need to solve large linear systems. We downsample shapes using a fast decimation algorithm, and use the coarse renders to optimize feature fields, making us highly robust to resolution (see Fig.~\ref{fig:timing}).

Based on these properties, we place our work in context with the relevant baselines in \cref{tbl:properties}. We emphasize that all existing works must re-solve an optimization problem for new handles (``New Handle w/o Optim."), and ours is the only method which affords both local and global control. 

\newcolumntype{C}[1]{>{\centering\arraybackslash}p{#1}} \newcommand{\cmark}{\textcolor{OliveGreen}{\ding{51}}}%
\newcommand{\xmark}{\textcolor{OrangeRed}{\ding{55}}}%
\newcommand{\indep}{\perp \!\!\! \perp}

\begin{table}
\centering
\label{tbl:properties}
\begin{tabular}{@{} l *{1}{C{1.3cm}} *{1}{C{1cm}}*{1}{C{1.1cm}}*{1}{C{1.5cm}} @{}}
\toprule
  & {\footnotesize\textbf{Global Semantics}} & {\footnotesize\textbf{Local Control}} & {\footnotesize\textbf{Size Robust}} & {\footnotesize\textbf{New Handle w/o Optim.}} \\ 
\midrule
  {\footnotesize\textbf{Classical}}\\
    {\footnotesize ARAP \cite{ARAP_modeling:2007}}  &  \xmark   & \cmark    &  \xmark & \xmark  \\
    {\footnotesize Biharmonic \cite{wang_linear_2015}}    &  \xmark & \cmark & \xmark & \xmark \\
  \midrule
   {\footnotesize\textbf{Neural}}\\
    {\footnotesize APAP \cite{yoo2024plausible}}  & \cmark & \xmark & \xmark & \xmark \\
    {\footnotesize DMH \cite{liu2021deepmetahandles}}   & \cmark & \xmark & \xmark & \xmark \\
  {\footnotesize NeuralMLS \cite{shechter2022neuralmls} }  & \xmark & \cmark & \cmark & \xmark \\
  \midrule[0.75pt]
  {\footnotesize\ourmethod{} (Ours)} & \cmark & \cmark & \cmark & \cmark \\
  \bottomrule                          
\end{tabular}
\caption{\textbf{Properties of Control-Based Deformation Methods.} ``Global Semantics" is whether the method can make visual-driven global deformations. ``Local Control" is the ability to smoothly deform the surface local to the transformed handle. ``Size Robust" is whether the weight computation scales robustly with mesh resolution. ``New Handle w/o Optim." is whether the method solves an optimization problem with every update to the control handle set. Our method (DFD) is the only work which incorporates these desiderata.}
\end{table}

\begin{figure*}
    \centering
    \begin{overpic}[width=\linewidth]{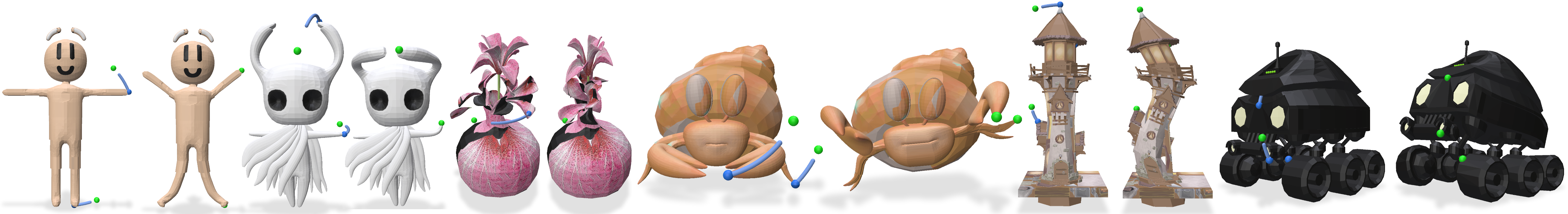}
    \put(7, -1){(a)}
    \put(21, -1){(b)}
    \put(34, -1){(c)}
    \put(51, -1){(d)}
    \put(70, -1){(e)}
    \put(87, -1){(f)}
    \end{overpic}
    \caption{\textbf{General Affine Transformations.} DFD weights effectively interpolate affine transformations to generate plausible pose changes. We can generate a variety of deformations by leveraging detected symmetries (a,b) (\ref{subsec:symmetry}) and localization control (b, d) (\ref{subsec:locality}).}
    \label{fig:affine}
\end{figure*}

\begin{figure*}
    \centering
    \includegraphics[width=\linewidth]{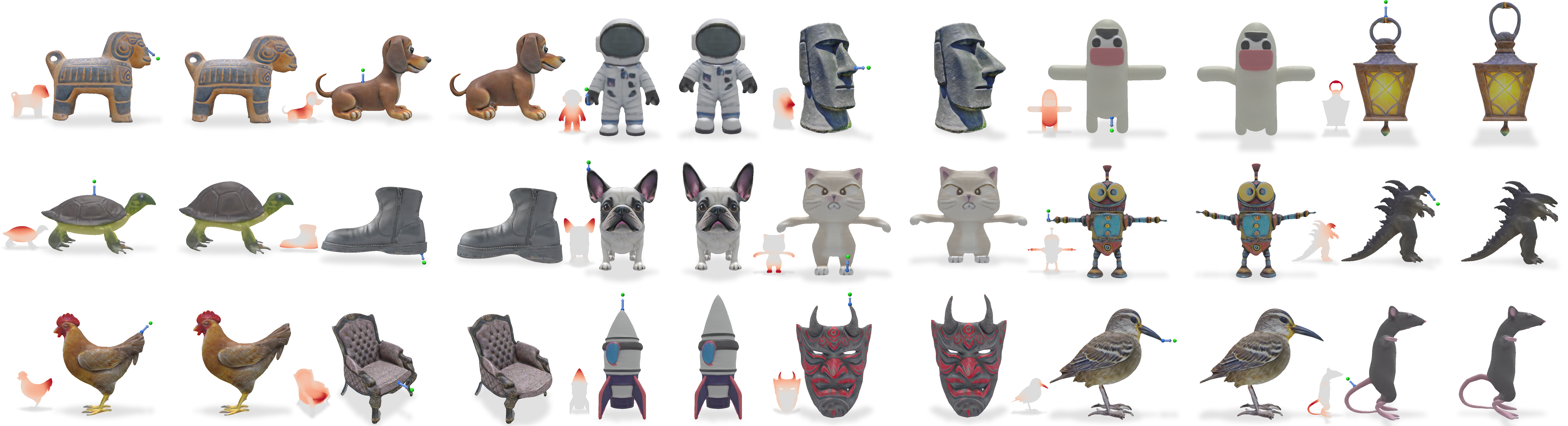}
    \caption{\textbf{Translation Edits.} Edit results from different handle (\textcolor{blue}{blue}) translations to target locations (\textcolor{green}{green}) using translations prescribed by APAP-Bench 3D. \citep{yoo2024plausible}. Weights are visualized as heatmap insets. A larger set of results from the dataset are shown in the supplemental.}
    \label{fig:gallery}
\end{figure*}

%% file: 03_method.tex
\section{Method} \label{sec:method}
DFD weights map control handle deformations to the rest of the shape through linear blending, enabling real-time and interactive mesh deformations (\ref{subsec:prelim}). We accelerate neural field optimization (pre-processing) through barycentric feature distillation, which maximizes the feature signal extracted from each render and allows efficient distillation of shapes with millions of elements (\ref{subsec:barydist}). Additionally, we demonstrate locality control, feature anchoring, and automatic symmetry evaluation capabilities (\ref{subsec:featureweighting}).

\subsection{Preliminaries}\label{subsec:prelim}
Given a mesh $\mathcal{M} = (V, F)$, with vertices $V \subseteq \mathbb{R}^3$ and faces $F$, our method predicts a DFD weight matrix $\mathcal{W} \in \mathbb{R}^{n \times n}$ ($n = |V|$) which defines the basis for a deformation subspace of the mesh. For a given set of deformations $\{D_1, \ldots, D_K\}$ (affine transformations assigned to vertices $v_{j_1}, \ldots, v_{j_k}$), we can compute the final position of vertex $i$ through an extended form of standard linear blending. 
\begin{equation}\label{eq:def}
    V'_i=(\max(1-\sum_{k=1}^K \mathcal{W}_{ij_k}, 0)D_0+\sum_{k=1}^K\mathcal{W}_{ij_k}D_k)V_i
\end{equation}
The term $\max(1 - \sum_{j=1}^K\hat{\mathcal{W}}_{ij}, 0) D_0$ represents a simulated control point which has some default transformation $D_0$. We assume in all our results that $D_0$ is the identity transformation, but can be set to some default affine transformation to be applied globally if desired. The max function is applied to avoid negative weights on the simulated control point, which can result in unintuitive behavior \cite{skinningcourse:2014}.

By including a simulated control point, we satisfy partition of unity for sparse handles (where $\sum_{k=1}^K\mathcal{W}_{ij_k} \leq 1$). This property is primarily enforced to guarantee no movement under identity transformation, and otherwise does not serve much practical purpose. Past work has demonstrated that dropping this constraint yields negligible changes to deformation quality (fig. 6 in \cite{BBW2011}). We also show in supplemental Fig.~\ref{supp:densehandles} that our method also produces reasonable results for dense handle configurations, where partition of unity is not guaranteed. Linear blending of affine transformations follows the same approach as past work \cite{BBW2011, wang_linear_2015}, which themselves follow a long tradition of linear averaging deformations in skeletal animation \cite{skinningcourse:2014}. 

\subsection{Barycentric Feature Distillation}
\label{subsec:barydist}

\begin{figure}
    \centering
    \includegraphics[width=0.9\linewidth]{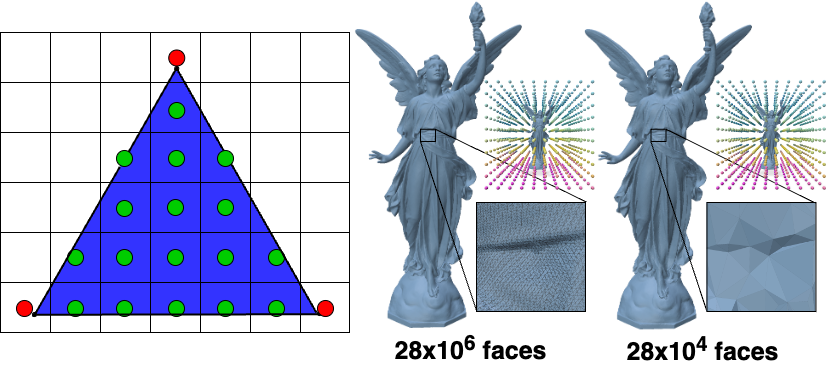}
    \caption{\textbf{Barycentric Feature Distillation.} (Left) Existing feature distillation methods use only \textcolor{red}{pixels} intersected by raster vertices. Barycentric distillation takes advantage of the known geometry to supervise with features at all \textcolor{green}{pixels} intersected by a triangle. (Right) High resolution meshes look visually unchanged even with extreme reduction using QEM (99\%). Consequently their feature fields are virtually identical (PCA insets). We opt to distill features using renders of low-resolution meshes, and use them to deform meshes at their original resolution.}
    \label{fig:barycentric}
\end{figure}

Barycentric feature distillation is motivated by the observation that existing work on feature distillation for surfaces waste much of the visual signal from renders \citep{Dutt_2024_CVPR, wimmer2024back}. \cref{fig:barycentric} (left) demonstrates this with a simple example of a rasterized triangle on the image plane. Current works distill features onto mesh vertices, which means only the pixels containing a vertex (red dots) contribute to features on the surface. Using a neural field and the triangle geometry, we can instead extract the 3D coordinates for all the pixels covered by the triangle (green), and optimize our neural field on the features from these pixels. Formally, we can use the rasterization process to define a function which assigns a 3D surface coordinate to each render pixel $(i,j)$.  
\begin{equation}
    P_{ij} = B(i,j)T(i,j)
\end{equation}
$T(i,j)$ is a matrix where the columns are the vertex positions of the triangle covering the center of pixel $(i,j)$ and $B(i,j)$ is a row vector containing the barycentric coordinates of the surface point at the pixel center. Note that $P_{ij}$ is only defined for pixels which have triangles intersecting the center. Let $Z_{ij}$ be encoded feature for pixel $(i,j)$. The training loss for our neural field $\Phi$ is
\begin{equation}
    \mathcal{L} = \sum_{(i,j) \in \Omega}\left\|\Phi(P_{ij}) - \frac{Z_{ij}}{||Z_{ij}||}\right\|^2
\end{equation}
where $\Omega$ contains the set of all pixels which have centers covered by a raster triangle. This method of supervising on features-per-pixel rather than features-per-vertex results in \emph{complete disentanglement} of the neural field sampling and the mesh resolution. Two meshes occupying the same volume in the field will induce the \emph{same sampling resolution}. 

Though rendering is typically fast, it becomes a non-trivial bottleneck for high resolution shapes. We observe that high-resolution meshes do not visually change much even with aggressive decimation, which motivates our decision to first downsample the shape with quadric error simplification (QEM) prior to rendering. We find empirically QEM to be substantially faster than rendering for shapes at most resolutions. In \cref{fig:barycentric} (right), a single render of the Lucy mesh (28 million faces) with Pytorch3D \cite{ravi2020pytorch3d} takes 5.7 minutes, whereas quadric mesh simplification with 99\% reduction and then rendering takes 3.7 seconds. Despite the reduction, the two meshes look visually identical. Distilling with barycentric features, however, is critical to ensuring the neural field distilled on the coarse mesh is effective for the mesh at its original resolution. We show in supplemental \cref{supp:baryabl} that standard vertex-based distillation on the coarse shape results in much worse DFD weights, even for the same number of training FLOPs. 

\begin{figure}
    \centering
    \includegraphics[width=\linewidth]{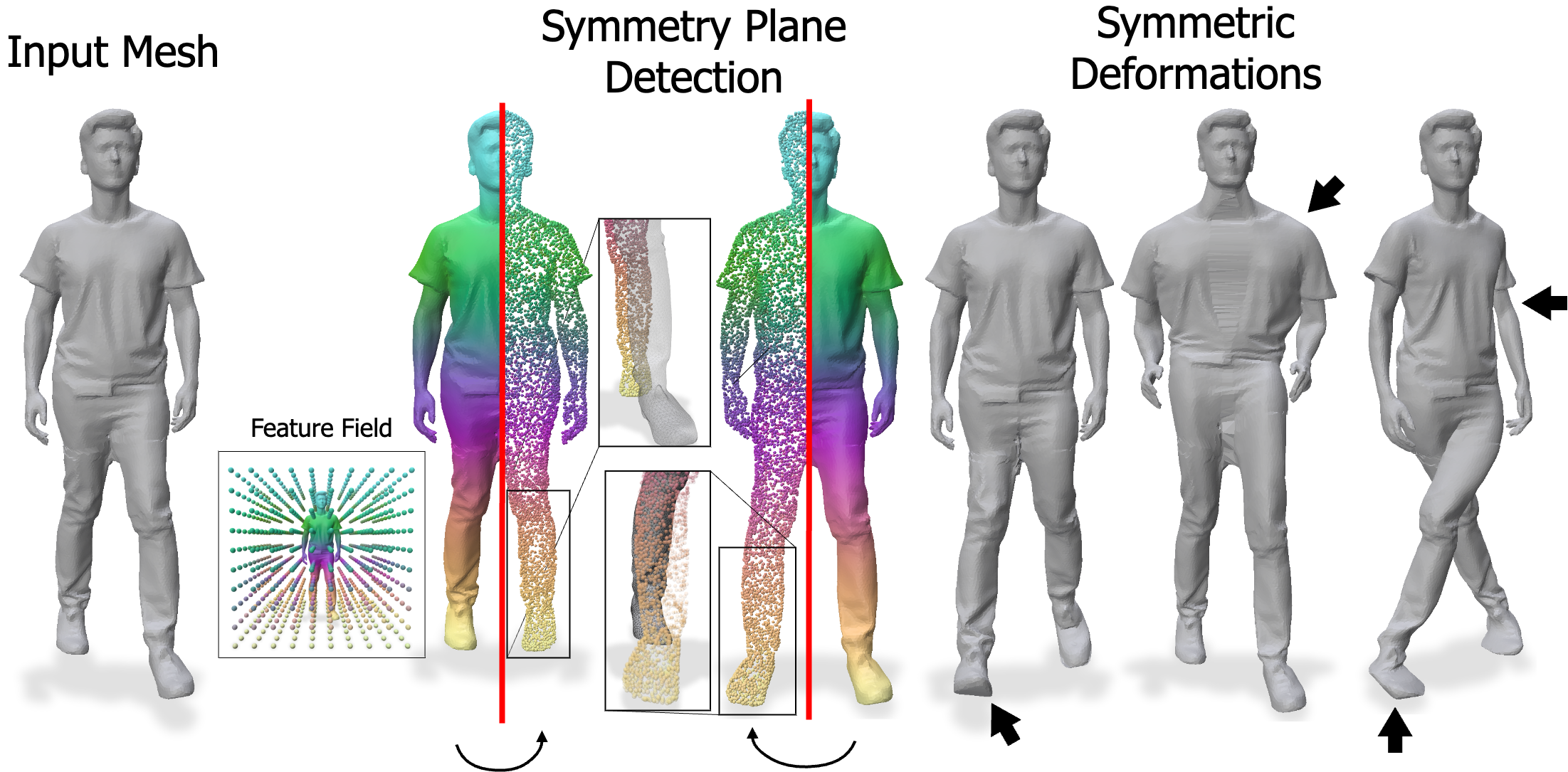}
    \caption{\textbf{Visual Symmetry Detection.} Our neural field representation returns visual features for arbitrary 3D points. This allows us to evaluate candidate symmetry planes on points  \emph{away from} the shape surface, which we use to identify symmetry planes where visual features are reflected. This identification is not constrained by extrinsic geometry or isometry constraints. Our symmetric deformations are generated by manipulating only \emph{one side of the shape.}}
    \label{fig:symmetry}
\end{figure}

\subsection{Feature Proximity Weighting}
\label{subsec:featureweighting}
Given a trained feature field $\Phi$ and mesh with vertices $V$, we have unit-norm distilled features $Z = \Phi(V)$. For some similarity function $F(x,y)$ which maps points $x$, $y$ to a similarity value between -1 and 1, our weights are defined as
\begin{equation}\label{eq:weight}
    \mathcal{W}_{ij} = \max(F(Z_i, Z_j), 0)
\end{equation}
$F$ can be any similarity function which falls within [-1, 1], and we find a simple $L2$-based function works well. 
\begin{equation}\label{eq:l2} 
F(Z_i, Z_j) = 1 - ||Z_i - Z_j||_2
\end{equation}
For unit norm features, $F$ is 1 when $Z_i$ and $Z_j$ are the same and -1 when the features are the furthest apart. Negative weights are undesirable \cite{BBW2011}, so we interpret all negative weights to represent unrelated features and clamp them to 0. For vertices $i,j$, the weight $W_{ij}$ is given by Eq.~\ref{eq:weight}.
Note that $Z$ is precomputed from a single forward pass of $\Phi$, so weights for new handles are obtained from computing $F$. Our choice of linear $F$ (\ref{eq:l2}) makes this weight calculation linear with respect to both vertex and handle count.  

\subsubsection{Feature Space Constraints} 
\label{subsec:latentanchor}
\begin{figure}
    \centering
    \includegraphics[width=0.8\linewidth]{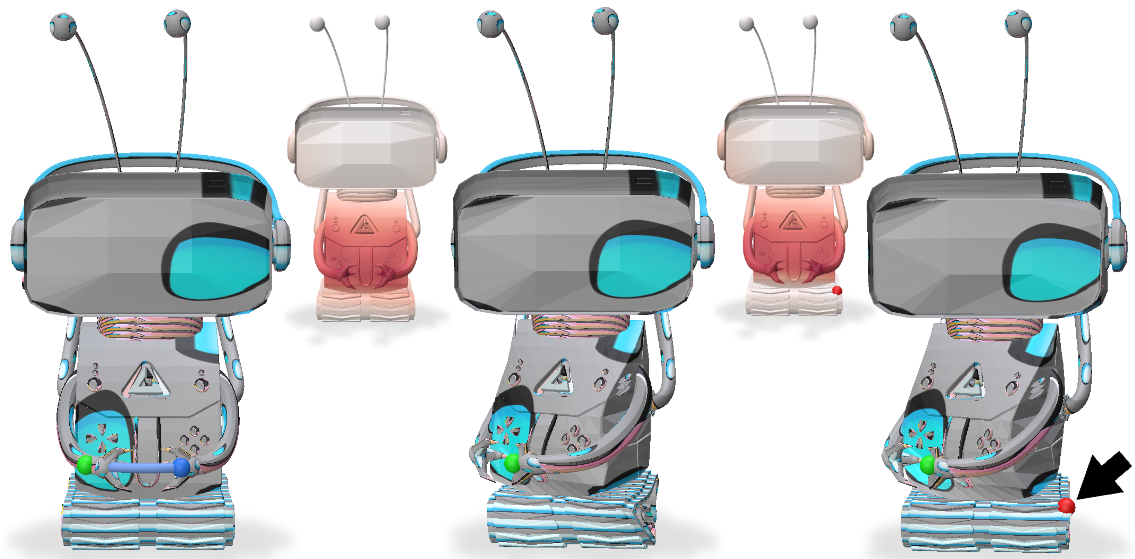}
    \caption{\textbf{Feature Space Constraints.} Fixed points in our framework constrain points with similar deep features. For example, we place a fixed point on the robot treads, which prevents it from twisting with the torso.}
    \label{fig:latentanchor}
\end{figure}

We propose a simple extension to our framework to account for point constraints. For fixed vertex indices $\{p_1, \ldots, p_k\}$, we can update $\mathcal{W}$ such that 
\begin{equation}
    \mathcal{W}_{ij} = \max(\mathcal{W}_{ij} - \max_{p_k}(W_{ip_k}), 0)
\end{equation}
Fixed points update the weights for vertex $i$ by subtracting the maximum weight between $i$ and the fixed points. This ensures $W_{ik} \leq 0$ for all fixed points $k$. The outer max ensures none of these adjusted weights become negative, so $W_{ik} = 0$. Practically, all points with similar visual features to the fixed points will be constrained. We demonstrate the effectiveness of these constraints in \cref{fig:latentanchor}.

\begin{figure}
    \centering
    \includegraphics[width=0.8\linewidth]{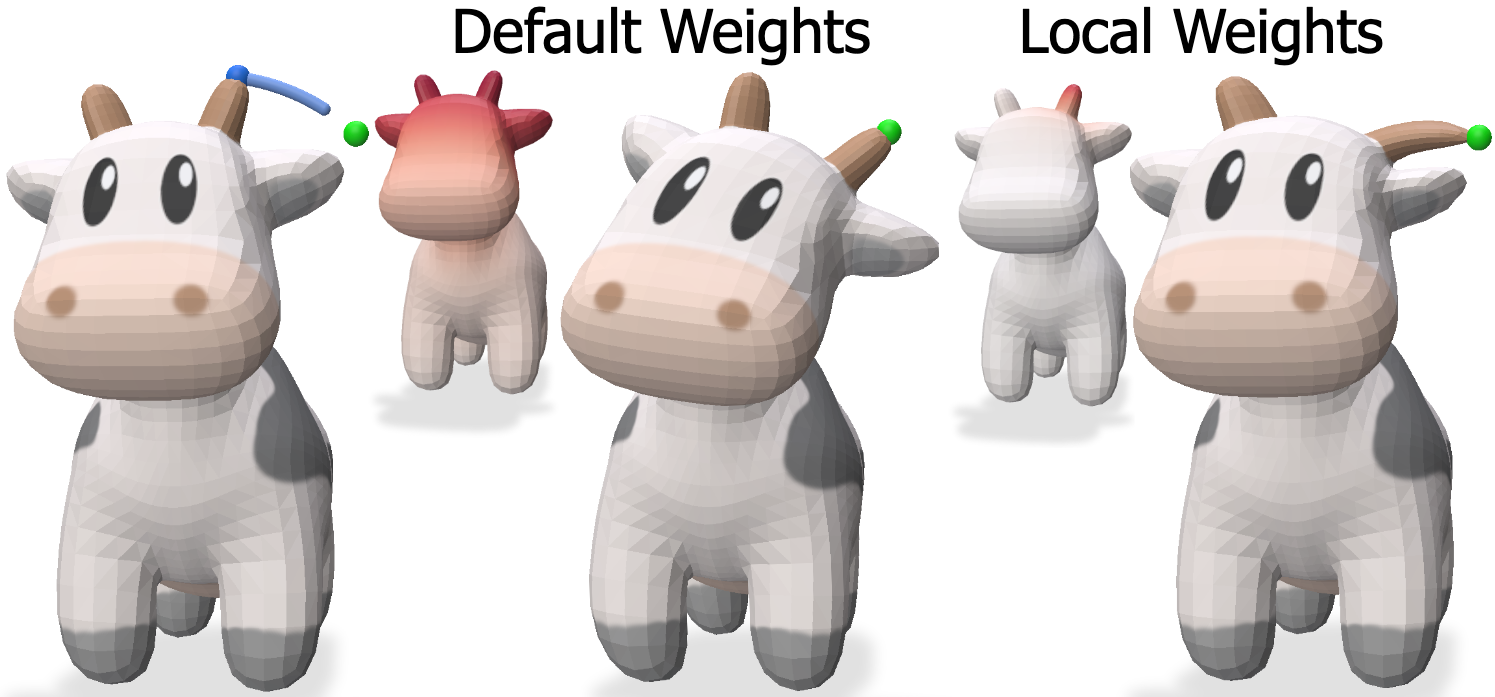}
    \caption{\textbf{Locality Weighting.} Locality weighting (\ref{eq:lambda}) allows precise control over the spatial extent of the target deformation. With default DFD weights, the deformation is a reasonable rotation of the cow head, and with locality weighting the same deformation instead bends just the horn.}
    \label{fig:geodesic}
\end{figure}

\subsubsection{Locality Weighting}
\label{subsec:locality}

A user may not always want deformations to result in global shape change. We introduce a user-defined parameter $\lambda$ which determines localization extent. If $\lambda > 0$, we update the weight matrix to localized weights $\mathcal{W}'$
\begin{equation}\label{eq:lambda}
    \mathcal{W}'_{ij} = \mathcal{W}_{ij} (1 - G_{ij})^\lambda 
\end{equation}
where $G_{ij}$ is the geodesic distance between vertex $i$ and vertex $j$ \cite{Sharp:2019:VHM} normalized such that the maximum geodesic distance is 1. As geodesic distance increases, the weights between points to the handle drops off to 0 with speed determined by $\lambda$. \cref{fig:geodesic} shows that by allowing user control over the dropoff, specific local features can be deformed.  

\subsubsection{Visual Symmetry Detection}
\label{subsec:symmetry}
Our neural field representation enables automatic evaluation of candidate symmetry planes. For a candidate plane $P$, let $V^+$ be the set of all mesh vertices on the positive side of the plane normal, and let $V^-$ be the vertices on the negative side. Let $R_P$ be the function which reflects points across $P$. We say there exists a visual symmetry along $P$ if 
\begin{align*}
    \frac{1}{|V|}(&\sum_i ||\Phi(V^+_i) - \Phi(R_P(V_i^+))||_2\\
    & + \sum_j ||\Phi(V^-_j) - \Phi(R_P(V_j^-))||_2) < \epsilon
\end{align*}

For symmetry-preserving deformations along plane $P$, we update the right-side term in \cref{eq:def} 
\begin{equation}\label{eq:symdef}
    \sum_{k=1}^K \mathcal{W}_{ij_k} \Rightarrow \sum_{k \in \Omega_i^+}\mathcal{W}_{j_ki}D_k + \sum_{k \in \Omega_i^-}\mathcal{W}_{j_ki}R_P(D_k)
\end{equation}
$\Omega_i^+$ is the set of handles on the same side of $P$ as vertex $i$, and vice-versa for $\Omega_i^-$. Handles in $\Omega_i^-$ have their deformations reflected across $P$ before being applied to $V_i$. 

Since similarity is computed based on visual features, the detected symmetries are not \emph{geometric}, but rather \emph{visual}. The closest related concept is intrinsic symmetry \cite{ovsjanikov_global_2008}, in which identical geometric parts in different poses are identified as symmetric. Our notion of visual symmetry takes this one step further, where the parts need not be intrinsically identical, but simply visually. 

\cref{fig:symmetry} shows an example of a shape which has visual but not extrinsic symmetry. Our neural field representation allows for evaluation of features that are not on the shape surface. In this example, we identify a vertical symmetry plane that enables symmetric deformations, such as switching the leg poses, broadening the shoulders, and even crossing the legs, by manipulating just one side of the shape. 

For all results shown, we evaluate symmetry planes spanning the primary axes and specify when the results apply detected symmetries. We set $\epsilon = 0.1$ for all results.

%% file: 04_experiments.tex
\section{Experiments} \label{sec:experiments}
\begin{figure*}
    \centering
    \includegraphics[width=\linewidth]{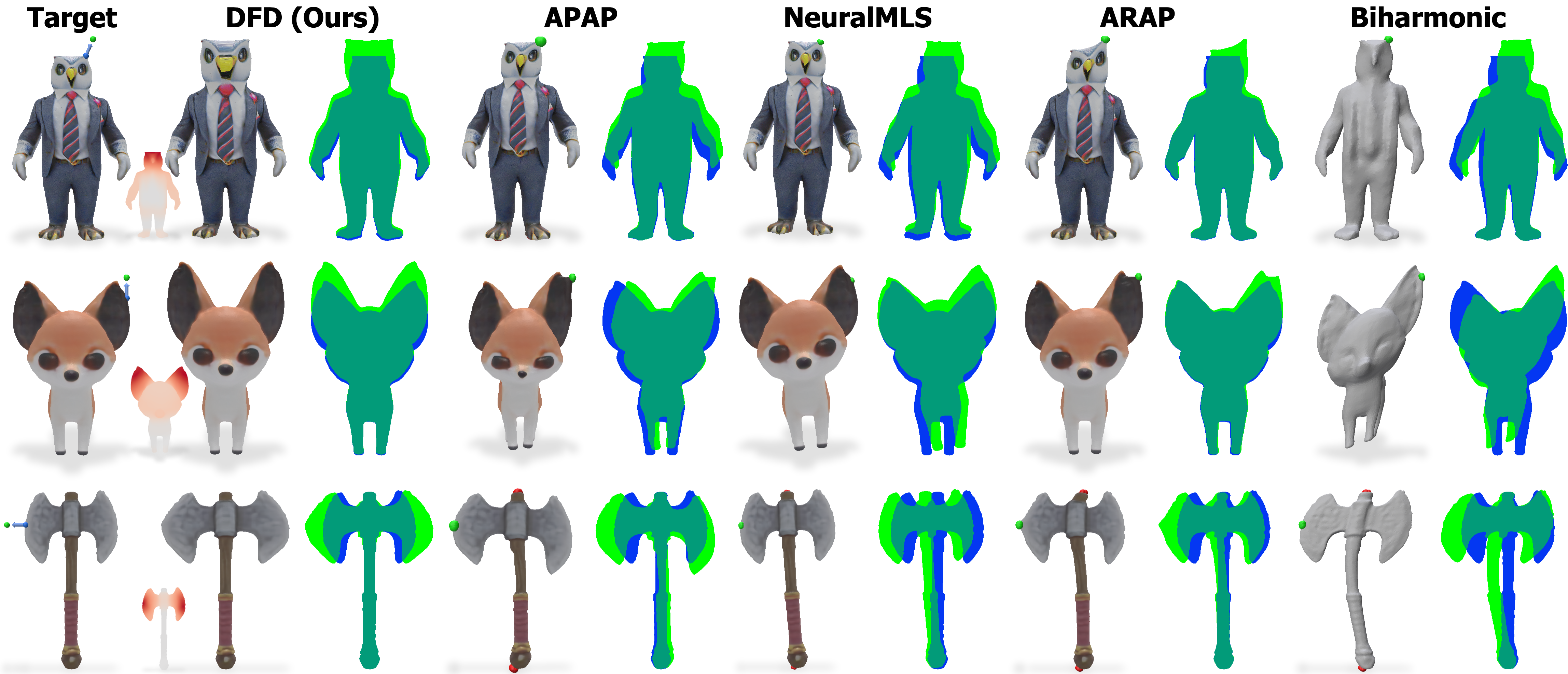}
    \caption{\textbf{APAP-Bench 3D Comparison.} Qualitative results using the shapes and handles shown in the APAP paper. DFD uses the single prescribed handle translation \emph{with no fixed points}. All other methods require many constraints to achieve reasonable deformations. These are generated with 0.01-ball sampling of handles/fixed points, following APAP. Control handle initial positions are in \textcolor{blue}{blue} and target positions in \textcolor{green}{green}. Fixed points are in \textcolor{red}{red}. Transparent silhouettes show deformation change, where the initial mesh is \textcolor{blue}{blue} and final mesh is \textcolor{green}{green}. DFD weights are shown in small heatmap insets. Biharmonic coordinates requires a tetrahedral mesh, whereby the texture is lost in conversion. Additional comparisons shown in supplemental figures \ref{supp:apapfull} and \ref{supp:apapsingle}, with and without 0.01-ball sampling, respectively.}
    \label{fig:apap}
\end{figure*} 
We evaluate against the baselines in \cref{tbl:properties}, and use the datasets from APAP (APAP-Bench 3D) \cite{yoo2024plausible} and DeepMetaHandles (DMH)~\cite{liu2021deepmetahandles} (1,363 shapes from the cars, tables, and chairs categories in ShapeNet~\cite{chang2015shapenet}). Some validation shapes are non-manifold (causing DMH, ARAP, and biharmonic coordinates to fail), so we generate manifold versions using~\cite{huang2018robust}. Our method is robust to non-manifoldness, as shown in supplemental \cref{supp:nonmanifold}. All DFD weights are distilled from DINOv2 \cite{oquab2024dinov2learningrobustvisual}. We QEM simplify shapes with over $50$k faces to $\leq$50k for barycentric feature distillation and do not simplify lower resolution shapes.

APAP-Bench 3D comes with prescribed handles and target positions. The DMH dataset comes with basis vectors and handles predicted by the trained DMH model. For our method, we use a single handle and target position by taking the largest-norm offset from the predicted basis.



Biharmonic coordinates require a tet mesh, so we tetrahedralize all shapes using FTetWild~\cite{ftetwild}. We correspond the original surface and the tet mesh with nearest neighbors, and we transfer the handles accordingly. All biharmonic results are visualized with the deformed tet mesh. Texture information is not transferred well, so we exclude textures.


\subsection{Qualitative Results}
\noindent\textbf{Affine Transformations.} DFD weights smoothly interpolate affine transformations while respecting shape semantics. We show affine deformations on shapes from Objaverse \cite{deitke2022objaverseuniverseannotated3d} in \cref{fig:affine}, which demonstrate the many axes of control within our framework. We show symmetric deformations enabled by our symmetry detection (a,b) (\ref{subsec:symmetry}), local deformations using locality weighting (b,d) (\ref{subsec:locality}), and pose changes using our base weights (a-c,d,f). 

\begin{figure*}[h]
    \centering
    \includegraphics[width=\linewidth]{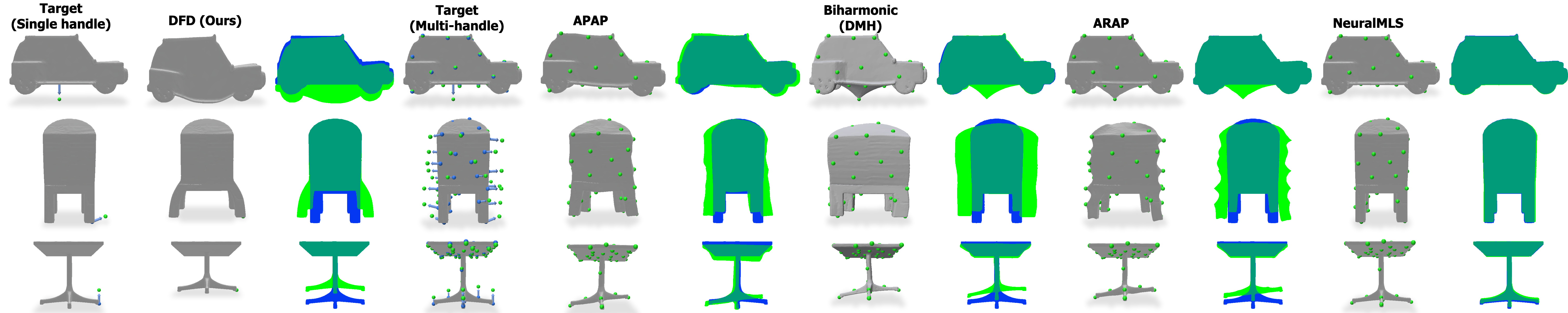}
    \caption{\textbf{DMH Comparison.} We compare against baselines using the DMH dataset \cite{liu2021deepmetahandles}. The baselines use 50 control handles with offsets predicted by DMH, while our method takes the single handle with highest-norm offset. DMH uses biharmonic coordinates as its deformation model, so they are equivalent. Our method generates deformations that are just as smooth as DMH and more visual/structure-aware.}
    \label{fig:shapenet}
\end{figure*}

\noindent\textbf{Translation-Based Editing.} \cref{fig:gallery} shows translation-based shape edits using DFD weights. These weights propagate edits to visually relevant features, such as the nose of the Moai sculpture, the arms of the robot, the chair cushion, and the horns on the demon mask. 

\noindent\textbf{Qualitative Comparisons.} \cref{fig:apap} compares the same shapes and handle transformations shown in APAP~\cite{yoo2024plausible} across the baselines. Our weights perfectly correspond key features on the shape, allowing for uniform stretching of the fox ears (row 2). Furthermore, symmetry detection allows us to generate uniform and symmetric scaling of the owl head and axe blades (rows 1,3). APAP does not consistently preserve symmetry due to its reliance on a noisy score distillation signal. Both ARAP and biharmonic coordinates are Laplacian-based, so the deformations are unsurprisingly non-visual and often result in global rotations/offsets due to poor placement of fixed points. Though handle sets in rows 1 and 2 are similarly placed (fixed points under the feet), the behavior of ARAP/biharmonic is inconsistent, demonstrating the difficulty in choosing performant handle sets. We show additional comparisons, with and without 0.01-sampling, in supplemental figures \ref{supp:apapfull} and \ref{supp:apapsingle}.

\cref{fig:shapenet} shows deformation comparisons on the DMH dataset. DMH uses biharmonic coordinates as its deformation framework, so they are synonymous in this comparison. We use handle sets and deformations predicted by DMH, making this a very strong baseline. Nevertheless, our method demonstrates smoothness on par with DMH with greater visual awareness. DMH will generate global rescaling of the chair, whereas our method can restrict the deformation to the legs (row 2). DMH will sometimes overconstrain (sharp artifacts in row 1) and lacks in symmetry awareness (uneven legs in row 3). Additional comparisons are in supplemental \cref{supp:dmh}. 

\subsection{Quantitative Results}
\noindent\textbf{Timing.} We conduct a timing analysis of DFD against the baselines to quantify our efficiency claims. To cover all methods, we measure timing of 3 phases (which may not all apply to each method): preprocess, bind, and pose. 
\begin{itemize}
    \item \textbf{Preprocess} time involves all the steps involved prior to computing the handle weights (bind). Biharmonic coordinates converts the surface mesh to a tetrahedral mesh, and prefactorizes the bilaplacian system for the linear solve. APAP renders the shape and finetunes a LORA model, and \ourmethod{} (our method) distills a feature field. 
    \item \textbf{Bind} time is the time taken to compute handle weights. Biharmonic solves a linear system over the tet elements, \ourmethod{} (our method) does a feedforward pass and feature distance calculation, and NeuralMLS trains a neural field. 
    \item \textbf{Pose} time involves computing the final deformed mesh. Both \ourmethod{} and biharmonic leverage the speed of linear blending, whereas ARAP solves a linear system, APAP conducts score distillation sampling \cite{poole2022dreamfusion}, and NeuralMLS solves a moving least squares problem.
\end{itemize}

\begin{figure}
    \includegraphics[width=\linewidth]{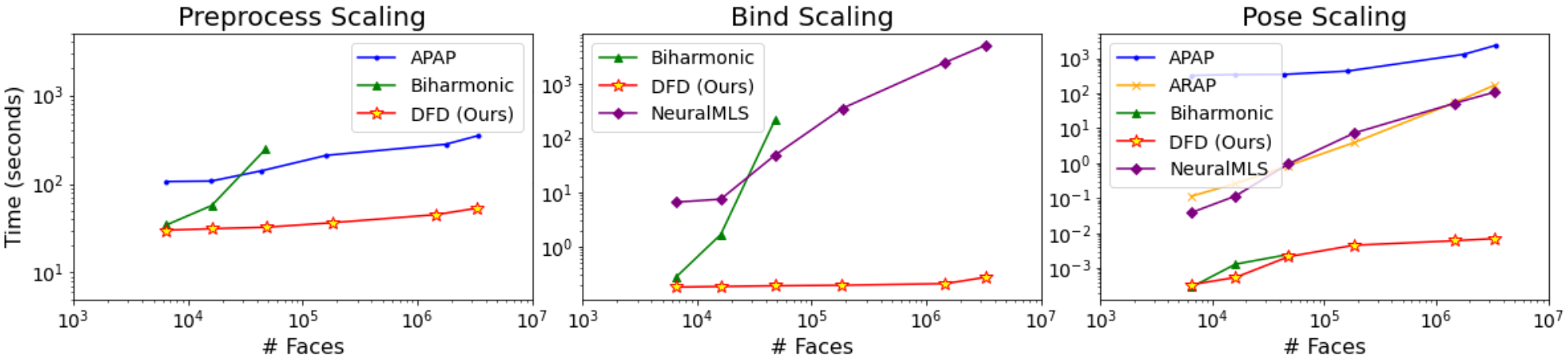}
    \caption{\textbf{Method Timing.} We compare timings across three stages (preprocess, bind, pose) for our datasets remeshed to different resolutions. Biharmonic coordinates fails in both the tetrahedralization preprocess and bind steps at resolutions higher than $10^5$ faces. Our method (DFD) is just as fast as biharmonic coordinates at the lowest resolution, and scales better than all methods across all phases. Our preprocess time is robust to mesh resolution due to barycentric feature distillation (\ref{subsec:barydist}). DFD also demonstrates sublinear scaling in both bind and pose time.}
    \label{fig:timing}
\end{figure}

We remesh each shape in our dataset to resolutions between $10^3$ to $10^7$ faces ($\sim$6,000 shapes) and compute method timings over them (\cref{fig:timing}, log scale). Biharmonic coordinates scale very poorly in the tetrahedralization and bind phases, and fails at resolutions past $10^5$ faces. Our method, on the other hand, demonstrates robust scaling for all phases across all resolutions. APAP also demonstrates good scaling but has a base runtime several orders of magnitude larger than DFD. ARAP and NeuralMLS exhibit both higher base runtimes and worse scaling than our method.

Though the analysis above already demonstrates our method's efficiency, it doesn't take into account the fact all other methods must re-optimize for new handles. OptCtrlPoints \cite{kim2023optctrlpoints} is a recent method that attempts to make the re-solve for biharmonic coordinates more efficient. We evaluate this accelerated re-solve against our method for new handle bind times in supplemental \cref{supp:optctrlpts} and show our method is still several orders of magnitude faster.

\noindent\textbf{User Study.} We conduct a user study using deformations from our evaluation datasets and 6 additional large handle deformations. To show our weights robustly interpolate both translations and rotations, we evaluate both translation-only (DFD-T) and affine (DFD-A) variants of our method. Users (N=37) selected the 2 ``most desirable'' deformations for each example, and we report the frequency each method is chosen in \cref{tbl:userstudy}. Both versions of our method are significantly preferred over the baselines (82\% for DFD-T and 79\% for DFD-A). Screenshots are in the supplemental \cref{supp:userstudy}. To measure perceptual realism, we conduct a second user study (N=23) which asks users to select the ``deformation which is most realistic and best preserves shape detail''. Our method is chosen by users 64\% of the time, ARAP 17.7\%, NeuralMLS 15.2\%, biharmonic 2.2\%, and APAP 0.93\%. Additional detail in \cref{supp:userstudy2}.

\begin{table}[h]
\small
\label{tbl:userstudy}
\begin{tabular}{ll|llll}
\toprule
\textbf{DFD-T} & \textbf{DFD-A} & ARAP & Biharmonic & APAP & NMLS \\ \hline
\textbf{82\%}  & \textbf{79\%}  & 19\% & 3\%        & 4\%  & 11\%     
\end{tabular}
\caption{\textbf{User Study.} We evaluate the translation (DFD-T) and affine (DFD-A) variants of DFD against baselines. Users (N=37) select the top 2 deformations for each example, and we report the frequency each method is chosen. NMLS stands for NeuralMLS.}
\end{table}

\subsection{Ablations}
\textbf{Barycentric feature distillation.} We ablate on barycentric distillation in supplemental \cref{supp:baryabl}. Specifically, we take the same approach as prior work and supervise the neural field solely on pixels which contain a vertex. We distill using the same decimated mesh and train for additional iterations to match the total \# FLOPs trained with under barycentric distillation. Despite this, the resulting DFD field produces weights which are neither smooth nor visual-aware on the original resolution shapes.

\noindent\textbf{Different image encoders.} We explore DFD weights extracted from different modern image models and find that they give surprisingly similar deformation results. Specifically we find that different image features tend to correlate the same structures, which indicates a convergence in semantic understanding of these different models. We show these results in supplemental \cref{supp:othermodels} and \cref{supp:imageinfluence}.

%% file: 05_conclusion.tex
\section{Conclusion} \label{sec:conclusion}
Deep Feature Deformation generates deformation weights using feature distances. These weights, without regularization, yield smooth and shape-preserving deformations. Barycentric feature distillation ensures our distillation is fast and resolution-agnostic linear blending enables interactive deformation, and the field representation allows visual symmetry detection. We expose classical axes of control through locality weighting and feature space constraints. Unlike prior methods we incorporate new handles without re-optimization, taking an important step towards true user-interactivity, which we demonstrate through a proof-of-concept GUI (supplemental videos). \\
\noindent\textbf{Limitations.} DFD weights are distilled on high resolution shapes in around a minute but still require per-shape optimization. Linear blending of extreme deformations has known issues (e.g. volume collapse) \cite{skinningcourse:2014} we do not resolve.

%% file: 06_acknowledgements.tex
\section{Acknowledgements} 
This project was funded by NSF 2402894, 2304481, the United States - Israel Binational Science Foundation (BSF) 2022363,
gifts from Adobe, Snap, Google, and The Bennett Family AI + Science Collaborative Research Program.

%% file: supplementary/supplementary.tex
\clearpage
\appendix
\setcounter{page}{1}
\maketitlesupplementary

\section{Ablations}
\label{supp:ablations}
\begin{figure}
    \centering
    \includegraphics[width=\linewidth]{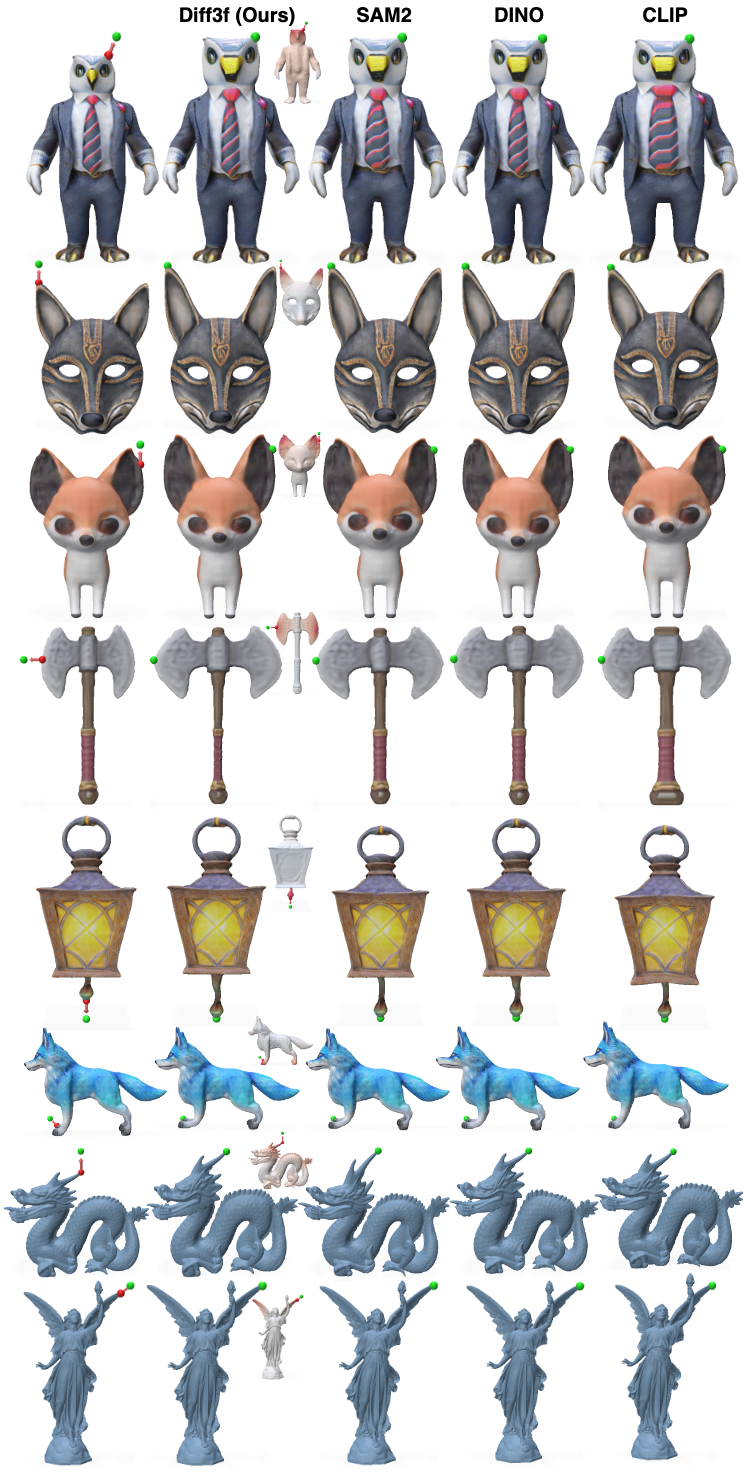}
    \caption{\textbf{Qualitative Results with Other Image Models.} We show deformation results using DFD weights computed using image features from other pretrained image models.}
    \label{supp:othermodels}
\end{figure}

\noindent\textbf{Different Image Encoders.} We show in \cref{supp:othermodels} additional deformation results using DFD weights computed using other image features, \emph{with no additional regularization or anchor points}. We find that deformation results are consistent and robust across all the image models we tested, though DINO and Diff3F give generally the best results.

\begin{figure}[h]
    \centering
    \includegraphics[width=\linewidth]{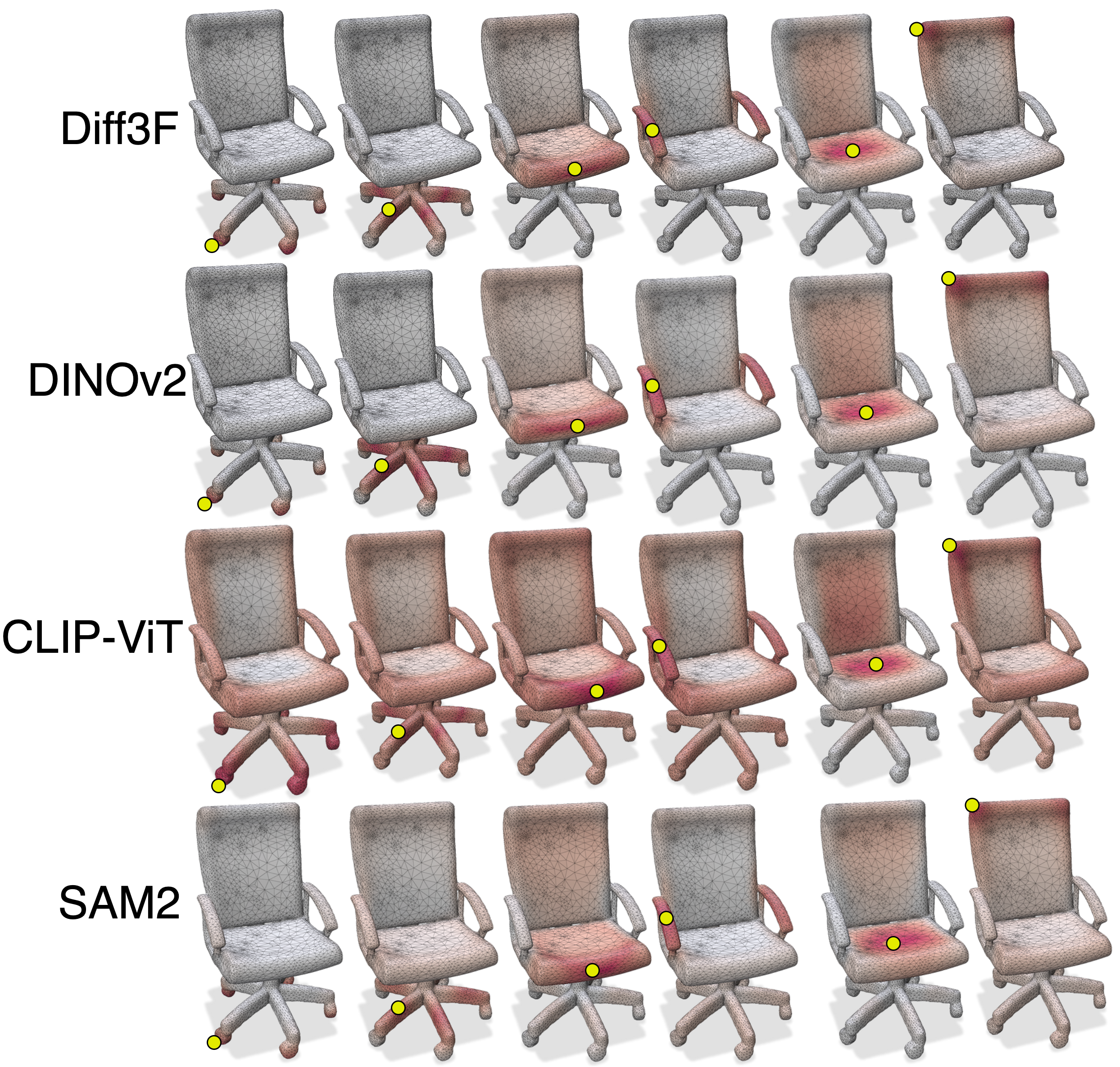}
    \caption{\textbf{Different Encoders.} By visualizing the DFD weights, we observe that pre-trained 2D foundation models contain approximately similar understanding of shape features and part relationships.}
    \label{supp:imageinfluence}
\end{figure}

Our weights also offer a unique perspective into image model interpretability. We visualize the DFD weights from different image models over the same shape in \cref{supp:imageinfluence}. We see that all 2D foundation models we tested converge to the same common global semantic understanding of shapes. We can also identify nuanced differences in image model behavior that coincides with prior reported observations. For example, CLIP-ViT \cite{radford2021learningtransferablevisualmodels} tends to focus much more on global understanding and less on local part relationships, whereas SAM2 \cite{ravi2024sam2} tends to better isolate local features. In practice, we find that DINO \cite{oquab2024dinov2learningrobustvisual} and Diff3F \cite{dutt2024diffusion} find the best balance between local and global shape understanding.

\begin{figure}[h]
    \centering
    \includegraphics[width=\linewidth]{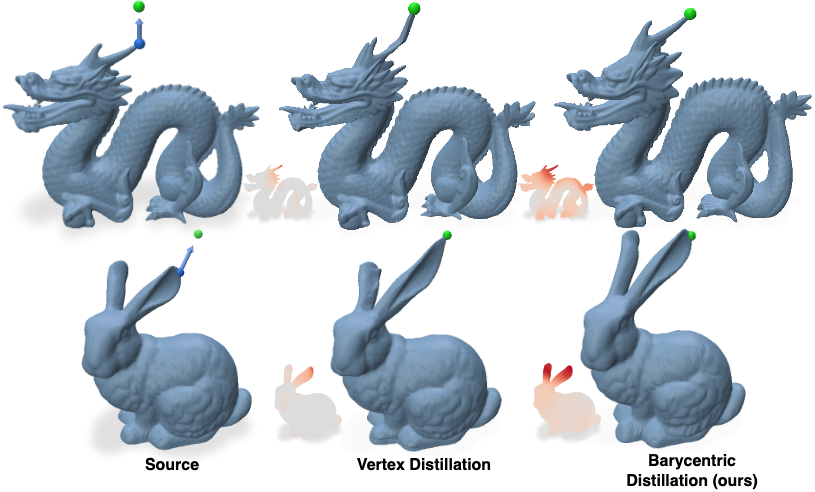}
    \caption{\textbf{Barycentric distillation ablation.} We distill DFD weights supervising only on vertex features on the decimated mesh and train for the same number of FLOPs as with barycentric distillation. The resulting deformations on the high resolution shapes are neither smooth nor visually-meaningful.}
    \label{supp:baryabl}
\end{figure}

\noindent\textbf{Barycentric Feature Distillation Ablation.} We ablate on barycentric feature distillation on high resolution shapes from the Stanford 3D scanning repository. We take each shape, simplify them using QEM decimation, and distill features into our feature field using either vertex distillation (the method used by prior works) or barycentric feature distillation. Vertex distillation takes only the features at render pixels which contain a vertex, whereas barycentric distillation makes use of every pixel which contains a point on the 3D surface. To ensure a fair comparison, we optimize the vertex distillation feature field for an equivalent \# FLOPs/pixel samples as our method. \cref{supp:baryabl} shows that without the dense field sampling offered by barycentric feature distillation, features distilled from coarse shape renders are unable to interpolate well to shapes at their original resolution. The deformations produces by the vertex distillation are neither smooth nor visually-meaningful. 

\begin{table}
\label{supp:optctrlpts}
\begin{tabular}{l|lll}
\toprule
Method / \# Control Points & 1  & 10 & 100 \\ \hline
\textbf{DFD (Ours)} & 0.004 & 0.012 & 0.034 \\
Biharmonic (OCP) & 4.6 & 11.35 & 21.8
\end{tabular}
\caption{\textbf{Rebind Time (s).} All existing methods require solving an optimization problem for every new set of control points, which is expensive (\cref{tbl:properties}). OptCtrlPoints (OCP) \cite{kim2023optctrlpoints} is a recent method which aims to make the re-solve more efficient. To compare rebind speeds, we take 20 random shapes from our dataset, precompute the OCP factorization, and randomly sample sets of 1, 10, and 100 control points 1000 times. Average time to rebind is reported in seconds. OCP is still limited by the optimization solve, and is $>$1000$\times$ slower than our method.}
\end{table}

\section{Rebinding Comparison to OptCtrlPoints}
\label{supp:rebind}
We quantitatively compare rebind times against OptCtrlPoints \cite{kim2023optctrlpoints}, a recent method for efficient rebinding through an updated solve to the biharmonic coordinates optimization problem. We take 20 random shapes from our dataset, precompute the OptCtrlPoints factorization, and then randomly sample sets of 1, 10, and 100 control handles. We report results in \cref{supp:optctrlpts}. Observe that for all control point set sizes, our method is still $\sim1000\times$ faster than OptCtrlPoints, demonstrating that bind time optimization is still a significant bottleneck. 

\begin{figure*}
    \centering
    \includegraphics[width=\linewidth]{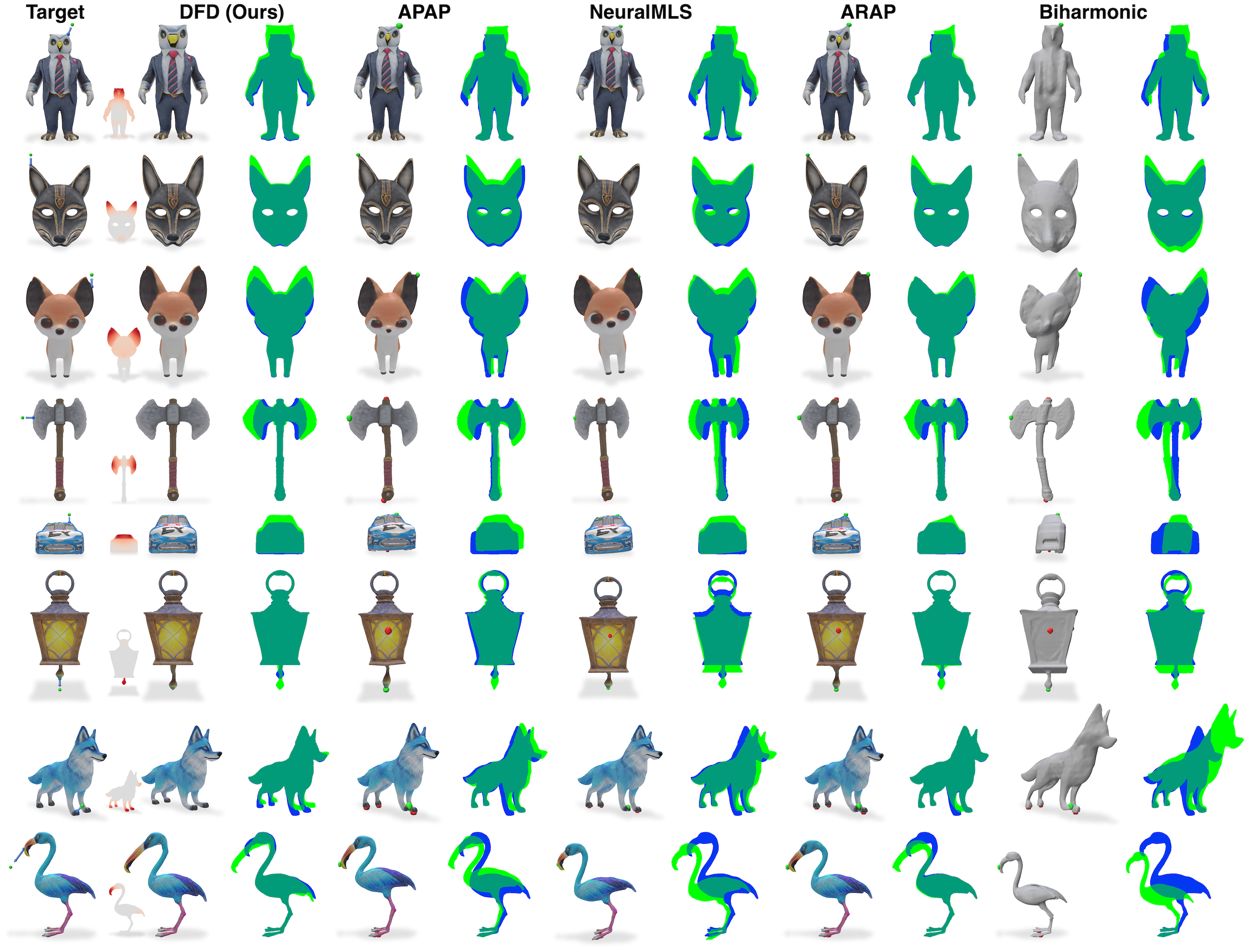}
    \caption{\textbf{Full Qualitative Comparison with APAP Results.} We compare against the full array of deformation results shown in the APAP \cite{yoo2024plausible} paper. Similar to the results in the main paper, in all examples our method produces visual and symmetry-aware deformations, whereas baselines produce undesirable global rigid transformations and general asymmetries.}
    \label{supp:apapfull}
\end{figure*}

\begin{figure*}
    \centering
    \includegraphics[width=\linewidth]{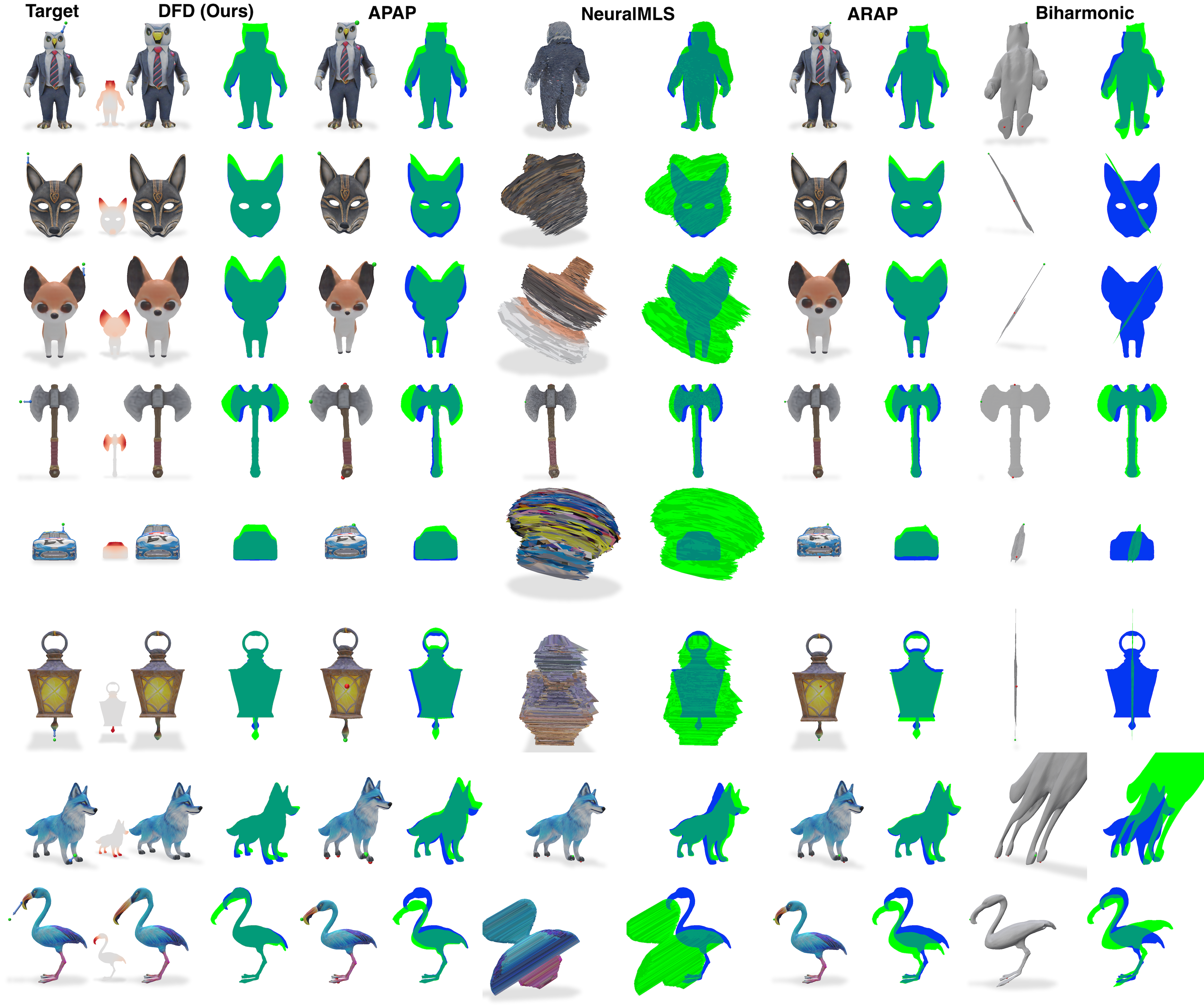}
    \caption{\textbf{Single-Handle APAP Comparison.} We compare against the baseline methods \emph{without} adding the 0.01-radius neighbors of each prescribed fixed point, which is a smoothing trick employed by APAP. Note that without the smoothing trick, some baselines fail completely (neuralmls, biharmonic) whereas other methods experience slightly worse artifacts (APAP, ARAP). Our method does not use fixed point constraints and does not rely on such smoothing hacks.}
    \label{supp:apapsingle}
\end{figure*}

\begin{figure*}
    \centering
    \includegraphics[width=\linewidth]{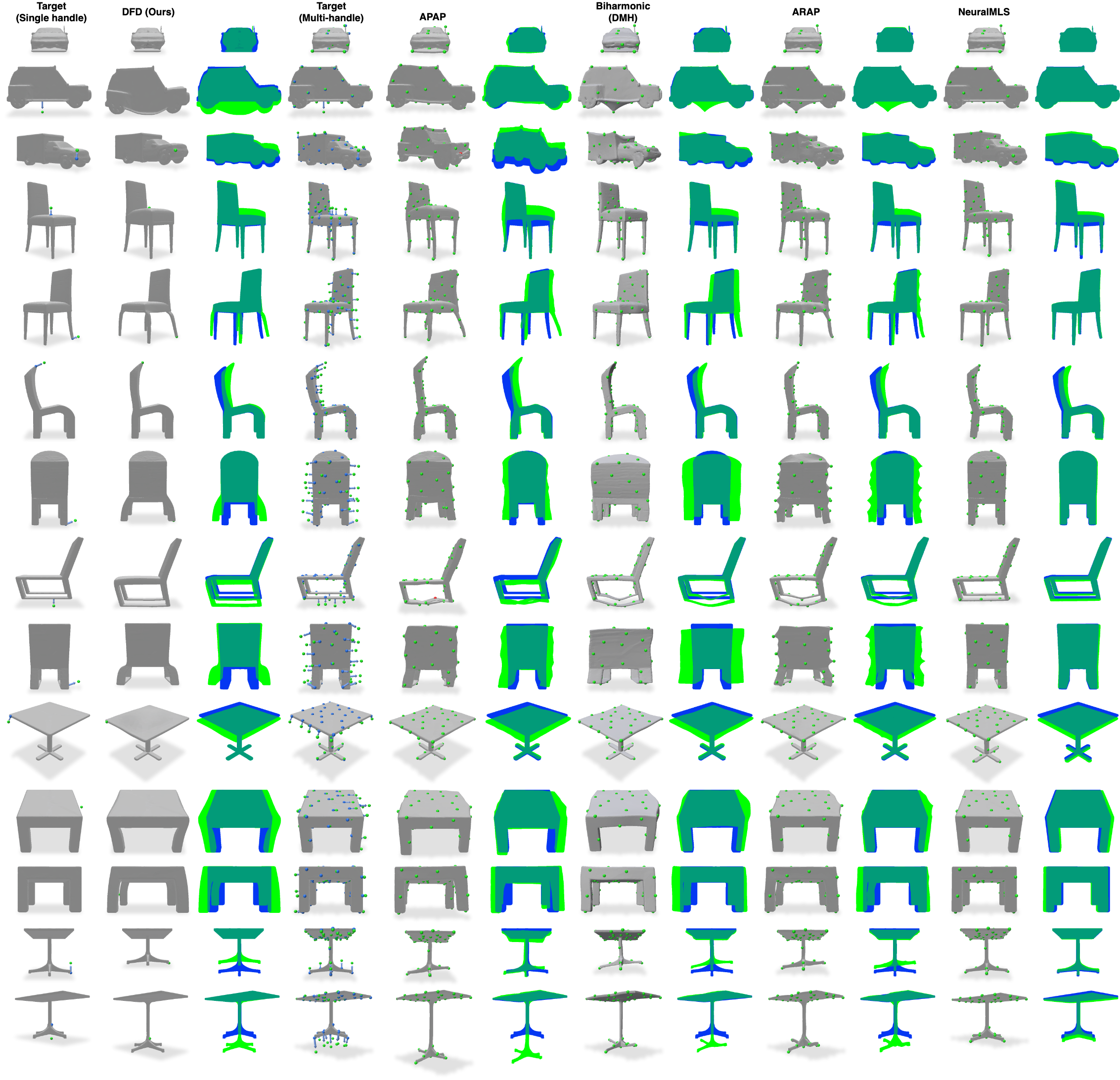}
    \caption{\textbf{Full DeepMetaHandles Comparison.} We show more examples comparing methods on the DMH dataset shapes and prescribed handle targets. As in the main paper, DFD weights are consistently just as smooth (or smoother) than DMH, while better-preserving shape semantics (e.g. part proportions, symmetries, etc).}
    \label{supp:dmh}
\end{figure*}

\section{Additional Comparisons to Baselines}
\label{supp:fullcomparison}
\noindent\textbf{APAP-Bench 3D.} We show in \cref{supp:apapfull} a full comparison of the baselines against DFD for all the deformations shown in the APAP paper \cite{yoo2024plausible}. We also show in \cref{supp:apapsingle} the full comparison without using the 0.01-ball sampling trick to increase the number of handle and fixed point constraints. As reported by APAP, all baselines other than APAP completely fail when dealing with a limited number of constraints. NeuralMLS and biharmonic coordinates degenerate, while ARAP produces global translations of the shape. APAP also experiences worse artifacts, whereas our method is completely stable with a single handle deformation. 

\noindent\textbf{DeepMetaHandles.} We show a larger set of qualitative comparisons on shapes from the DeepMetaHandles dataset in \cref{supp:dmh}. As observed in the main paper, DFD weights produce deformations which are consistently smoother and more symmetry/part preserving than the baseline methods. DMH is the strongest baseline here because the model is both trained on this data and used to predict the handle deformations. Despite this, it is still inconsistent in predicting smooth and visual-aware deformations (e.g. rows 2-4, 8,9,11,13). 

\begin{figure}
    \centering
    \includegraphics[width=\linewidth]{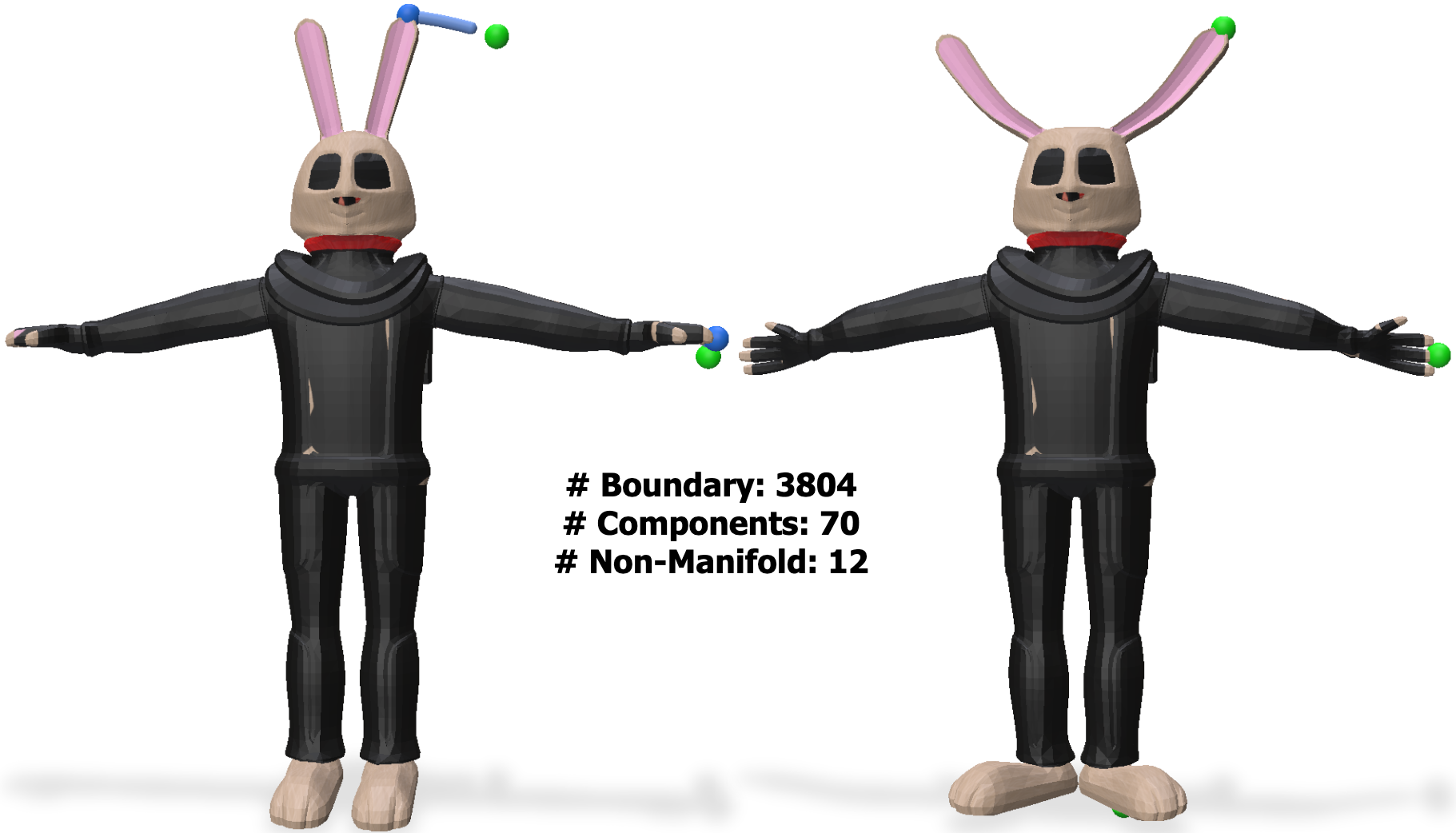}
    \caption{\textbf{Topology Robustness.} DFD weights are highly robust to topological defects. The example shown has 3,804 boundary edges, 70 disconnected components, and 12 non-manifold edges, but our weights still generate high quality deformations.}
    \label{supp:nonmanifold}
\end{figure}

\section{Topology Robustness}
DFD weights are extremely robust to topological defects, thanks to the visual nature of the feature field supervision. We demonstrate this by showing pose changes on a problematic model in \cref{supp:nonmanifold}. The example shown has 3,804 boundary edges, 70 disconnected components, and 12 non-manifold edges, but our weights are able to still smoothly interpolate deformations and produce plausible pose changes. 

\begin{figure}[h]
    \centering
    \includegraphics[width=\linewidth]{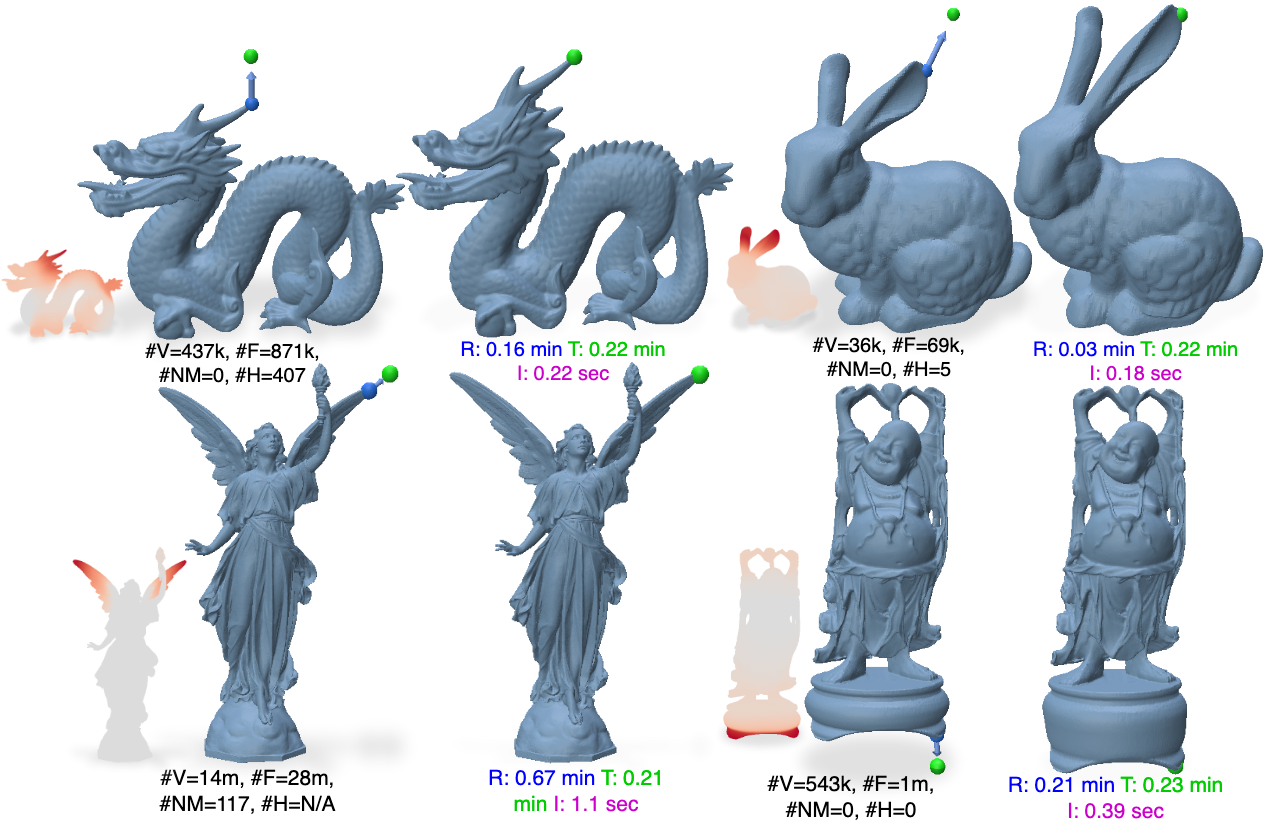}
    \caption{\textbf{High Resolution Deformation.} We test our distillation approach on very large meshes from the Stanford 3D Scanning Repository. \#V is the number of vertices, \#F is the number of faces, \#NM is total number of non-manifold elements, \#H is the number of holes. \textcolor{blue}{R} reports the decimation/rendering time during distillation. \textcolor{green}{T} reports the remainder of the time taken for distillation. \textcolor{violet}{I} reports the inference time. The mesh resolution influences the rendering stage (\textcolor{blue}{R}), but otherwise distillation time (\textcolor{green}{T}) is \emph{independent} of the mesh resolution.}
    \label{supp:stanford}
\end{figure}

\section{High Resolution Distillation}
We visualize distillation timings and deformations on very high resolution shapes from the Stanford 3D Scanning Repository in \cref{supp:stanford}. Because these shapes are generated from scans, they contain topological defects, which are reported under each shape (\#NM reports number of non-manifold elements and \#H is the number of holes). We furthermore report the QEM decimation and rendering time (\textcolor{blue}{R}), the feature field distillation time (\textcolor{green}{T}), and the pose/inference time (\textcolor{violet}{I}). We emphasize that the feature field distillation itself is completely agnostic to mesh resolution, which is why \textcolor{green}{T} is largely constant across all shapes. Pose time \textcolor{blue}{I} does scale with shape resolution, especially when memory limits require batching of the feedforward pass (Lucy model), but it is still very fast. The majority of the bottleneck is limited to \textcolor{blue}{R}, which scales robustly with mesh resolution thanks to the efficiency of QEM relative to rendering. Even at the highest resolution, the entire distillation process end-to-end is under a minute.

\begin{figure}[h]
    \centering
    \includegraphics[width=\linewidth]{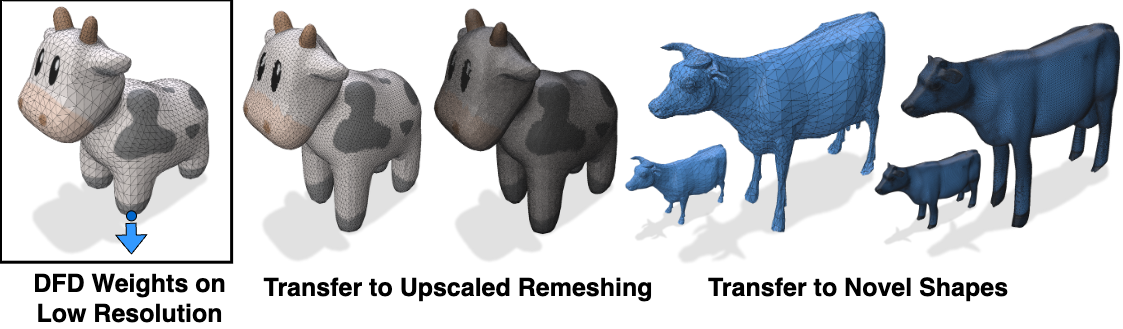}
    \caption{\textbf{Weight Generalization.} Distilling features into a neural field allows the same DFD weights to be applied to arbitrary resolution remeshings or even novel instances of the shape class.}
    \label{supp:remesh}
\end{figure}

\begin{figure}[h]
    \centering
    \includegraphics[width=\linewidth]{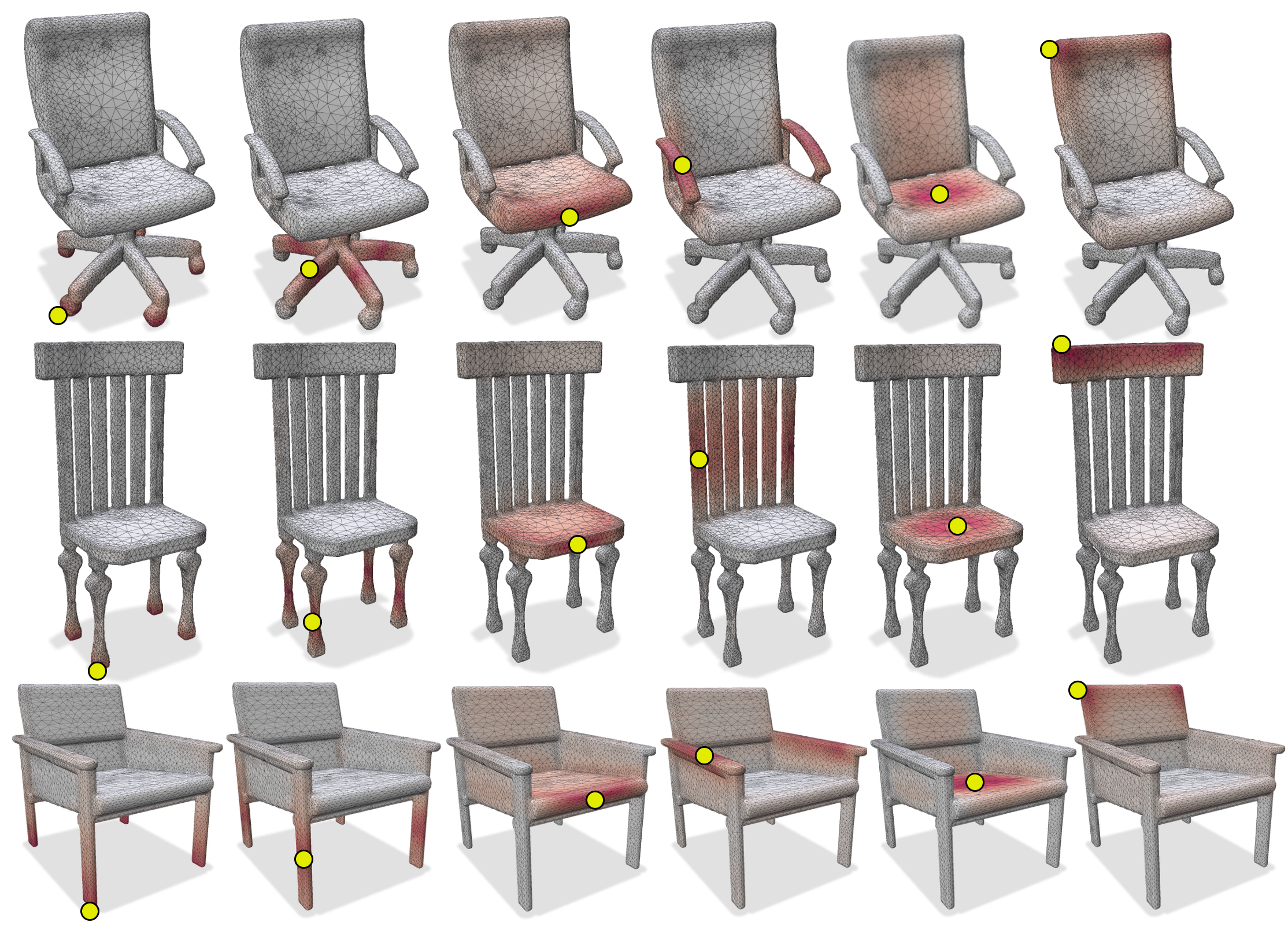}
    \caption{\textbf{Consistent Weights.} Our DFD weights identify consistent visual relationships across shapes within the same class.}
    \label{supp:multiinfluence}
\end{figure}

\section{Weight Generalizability}
\noindent\textbf{Novel Shape Instances/Remeshings.} Thanks to barycentric feature distillation, our weights generalize well from coarse shapes to higher-resolution remeshings. Furthermore, our distilled feature field can even be used to deform novel shapes within the same shape class, as shown in \cref{supp:remesh}. The neural field representation allows for any point in the ambient space to receive a visual feature, and thanks to the smooth nature of the field, novel shapes with similar visual parts in similar spatial regions can share the same field. Reusing these weights allows for similar visual-aware deformations, such as the co-deformation of the cow legs. 

\noindent\textbf{Consistent Shape Understanding.} In \cref{supp:multiinfluence}, we show that even across widely varying geometries, shapes within the same class (e.g. chairs), will induce DFD weights which identify similar visual relationships, such as the legs, armrests, backrest, and seat of the chair models. 

\section{Dense Handle Results}
We evaluate our method's performance on dense handle configurations with the same handles used by the baselines in Fig.~\ref{fig:shapenet}. Observe that even though partition of unity is no longer guaranteed under dense handles, our method's results are still reasonable and symmetry preserving.

\begin{figure}[h]
    \centering
    \includegraphics[width=\linewidth]{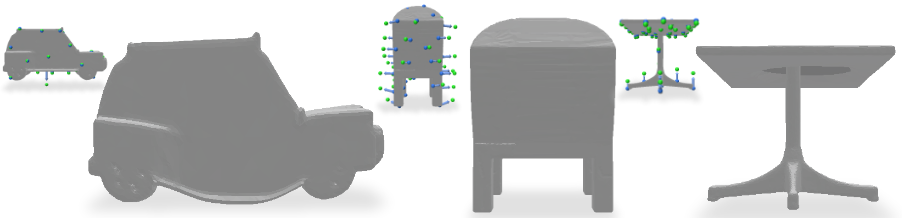}
    \caption{\textbf{Dense Handle Results.} We show deformations using our method using the same dense handle configurations as the baselines in Fig.~\ref{fig:shapenet}.}
    \label{supp:densehandles}
\end{figure}

\section{Surface Metrics}
\begin{figure}[h]
    \centering
    \includegraphics[width=\linewidth]{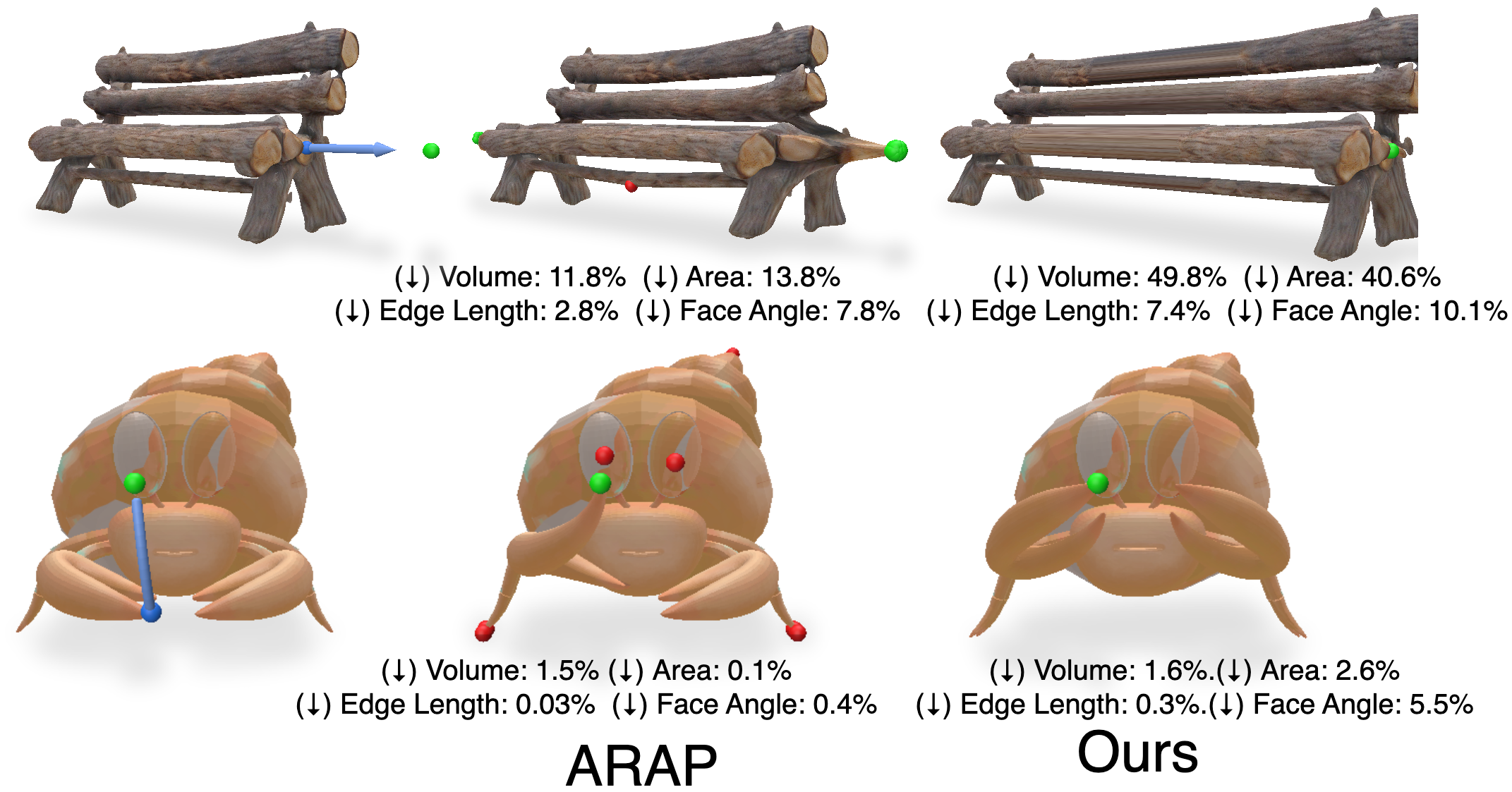}
    \caption{\textbf{Surface Metrics.} Though our method produces more surface distortion in terms of traditional metrics, the visual results demonstrate low surface distortion does not necessarily translate to a more realistic or natural deformation.}
    \label{supp:metrics}
\end{figure}

We report the same surface metrics from Implicit-ARAP \cite{baieri2025implicit} for two example deformations in \cref{supp:metrics}. Though our method produces greater distortion along all four metrics, we observe that visually our results maintain greater realism based on the shape's semantics and part being deformed. 

\section{Interactive GUI}
We provide video examples in the \verb|examples| folder of our interactive GUI demonstrating the semantic understanding and interactivity enabled by DFD weights. 

\section{User Study Screenshots.} 
\begin{figure*}
    \centering
    \includegraphics[width=\linewidth]{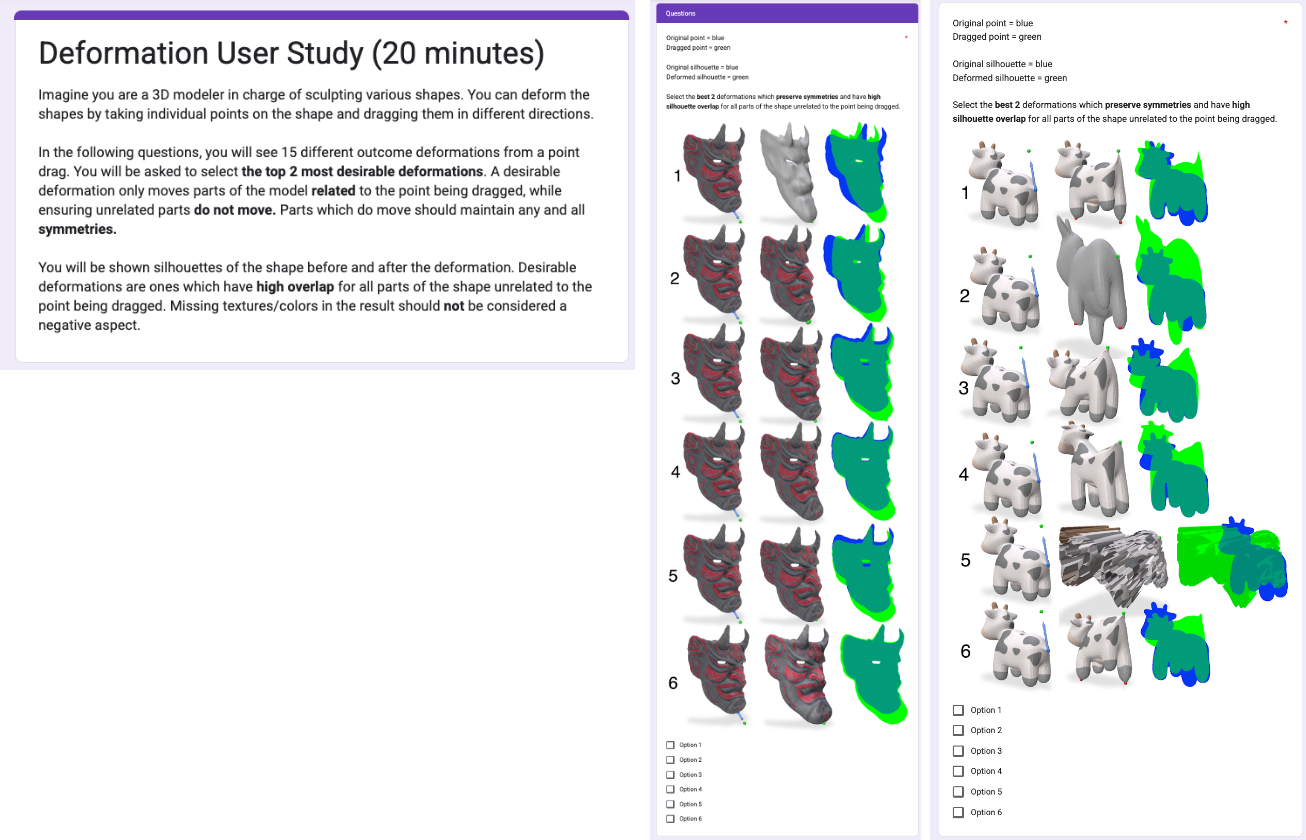}
    \caption{\textbf{User Study.} We show screenshots from our user study comparing deformations for various examples shown in the paper across all the methods (including both the affine and translation variants of our method).}
    \label{supp:userstudy}
\end{figure*}

We show screenshots from our user study in \cref{supp:userstudy}. We instruct users to select the top 2 deformation results for 15 different shape deformations from our datasets. We anonymize the 6 different methods (which include both the translation and affine variants of our results) and randomly shuffle their presentation order. We collect responses from 37 users. \cref{tbl:userstudy} shows that both variants of our method are highly preferred relative to the other baselines. 

\section{Realism User Study.}
\label{supp:userstudy2}
\begin{figure}
    \centering
    \includegraphics[width=0.49\linewidth]{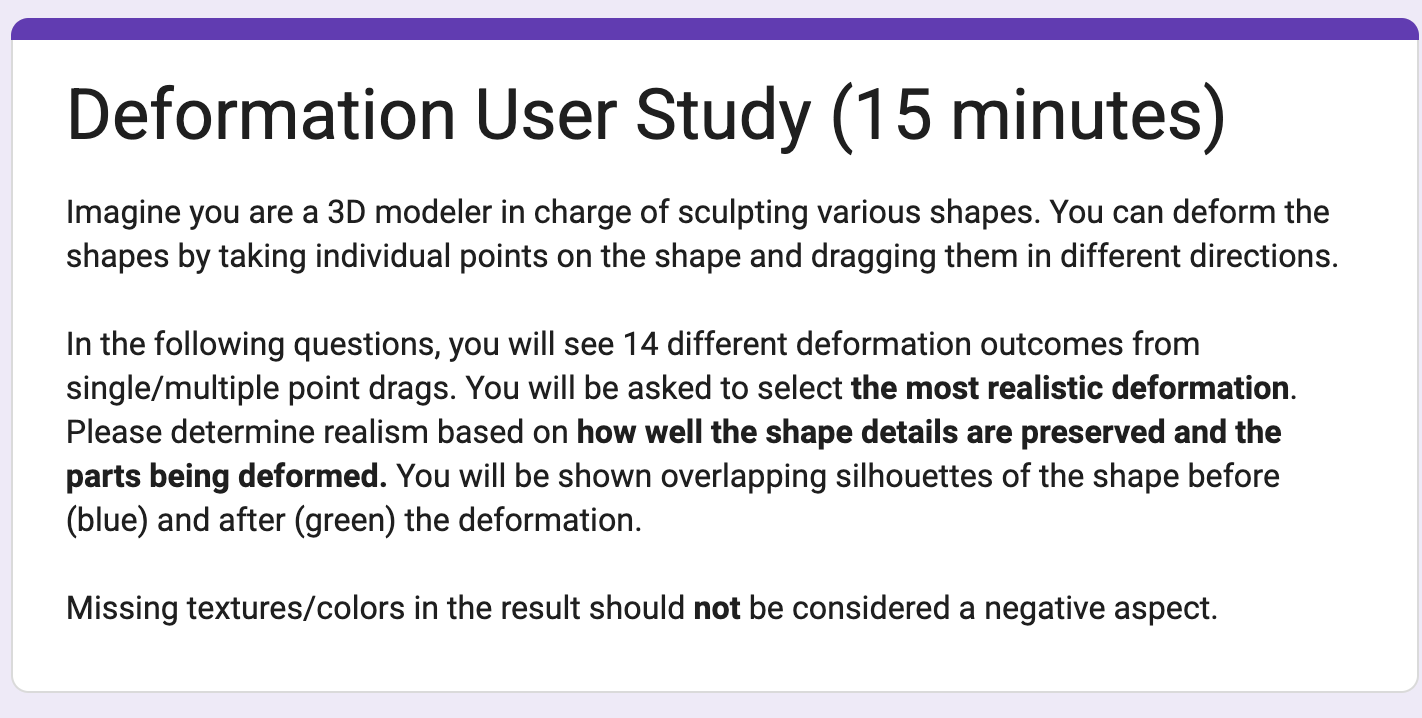}
    \includegraphics[width=0.49\linewidth]{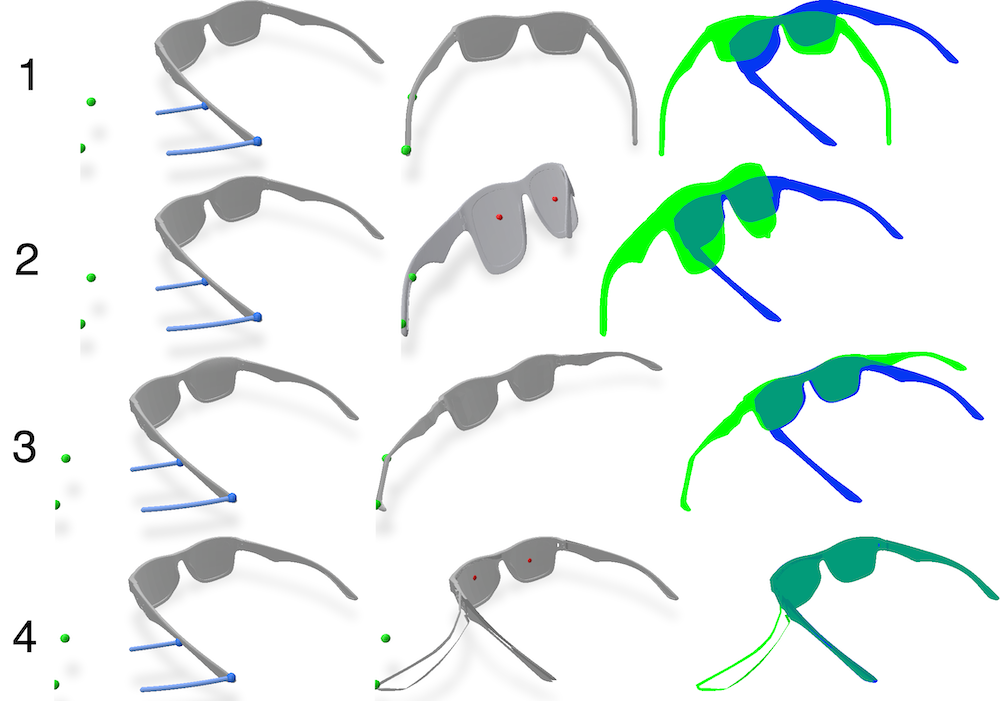}
    \caption{\textbf{Realism User Study.} We show screenshots from our second user study, which evaluates the quality of the deformations on the basis of realism.}
    \label{supp:userstudy2_screenshots}
\end{figure}

In order to focus on realism evaluation, we conduct a second user study asking users to select the ``more realistic deformation'' among the methods (N=23). We show screenshots in \cref{supp:userstudy2_screenshots}. Our
025 method is chosen by users 64\% of the time, ARAP 17.7\%, NeuralMLS 15.2\%, biharmonic 2.2\%, and APAP 0.93\%, demonstrating our results are more reliastic and shape-preserving from a user perspective. 

\section{Technical Details}
\label{supp:technicaldetails}
Our feature field is parameterized by a 4-layer MLP with ReLU non-linearities and a LayerNorm after each hidden layer. The output is normalized to unit norm. 

For all experiments, we sample 24 views using Fibonacci sampling. We optimize our feature field for 15 iterations, and render at $512 \times 512$. We use the Fast-Quadric-Mesh-Simplication~\cite{garland1997surface,sullivan2019pyvista} wrapper from PyVista to perform our decimation. Our qualitative results use DINO features, though all image models we tried gave reasonable results (see supplemental). All experiments are run on a single A40 GPU with 48GB RAM. 